\documentclass{article}

\usepackage[margin=1in]{geometry}

\usepackage[utf8]{inputenc}
\usepackage[T1]{fontenc}

\usepackage{natbib}
\usepackage{xurl}
\usepackage{hyperref}
\usepackage{booktabs}
\usepackage{caption}
\usepackage{subcaption}
\usepackage{pdflscape}
\usepackage{abstract} % control abstract margins
\usepackage{adjustbox}
\usepackage{amsmath,amssymb,amsfonts}
\usepackage{amsthm}
\usepackage{enumitem}
\usepackage{nicefrac}
\usepackage{microtype}
\usepackage[table]{xcolor}
\usepackage{mdframed}
\usepackage{wrapfig}
\usepackage{etoolbox}
\AtBeginEnvironment{abstract}{\setlength{\parindent}{0pt}}

\newmdtheoremenv[
  linewidth=0.8pt,
  roundcorner=3pt,
  backgroundcolor=gray!6,
  innertopmargin=0.6\baselineskip,
  innerbottommargin=0.6\baselineskip,
  innerleftmargin=0.8em,
  innerrightmargin=0.8em
]{boxedproposition}[proposition]{Proposition}

\usepackage{graphicx}
\usepackage[percent]{overpic}
\usepackage{tabularx}
\usepackage{longtable}
\usepackage{float}
\usepackage{placeins}
\usepackage{multirow}
\usepackage{rotating}
\usepackage[ruled,vlined,linesnumbered]{algorithm2e}

\hypersetup{colorlinks=false, pdfborder={0 0 0}}

\newcommand{\ours}{\textsc{CoRe}}

\title{Correction-space Cross-variate Interaction for Test-time Adaptation in Time Series Forecasting}

\author{
\begin{tabular}{c}
Yuanyuan Deng \qquad Mykola Pechenizkiy \qquad Songgaojun Deng\\[0.6em]
\small Eindhoven University of Technology\\
\small Eindhoven, The Netherlands
\end{tabular}
}

\date{}

\begin{document}

\maketitle

\begin{abstract}
Test-time adaptation (TTA) is a promising paradigm for handling distribution shift in time-series forecasting (TSF), where models adapt at inference time, often leveraging delayed observed data to refine predictions.
In the multivariate setting, distribution shifts often exhibit cross-variate dependencies, yet existing TSF-TTA methods adapt each variate independently and ignore this cross-variate structure.
Exploiting such structure motivates cross-variate interaction, but coupling variates through backbone predictions introduces direct pathways for mixing uncorrected errors across variates, a concern under the delayed supervision of TSF-TTA.
We identify the \emph{interaction space} as a key design choice, and show that acting on adapter corrections that refine backbone outputs, the \emph{correction space}, rather than on the predictions themselves, avoids directly propagating backbone errors across variates.
We build on this to propose \textsc{CoRe} (\textsc{Co}rrection-space Interaction \textsc{Re}finement),  realizing correction-space interaction through (i) Shared-anchor Correction Refinement (SCR), which combines each variate's correction with a shared anchor through a parameter-efficient bottleneck, and (ii) input-conditioned spectral gating, which adaptively modulates the refinement from the current input window.
Across seven backbones, six datasets, and four prediction horizons,
\textsc{CoRe} reduces MSE by 25.82\% on average over backbones and 10.57\% over the state-of-the-art TSF-TTA method,
with stronger gains at medium-to-long horizons and modest computational overhead. Data and code are available at: \url{https://github.com/yyddou/CoReTTA}.
\end{abstract}

\section{Introduction}

Time-series forecasting (TSF) supports decision-making across domains
including energy, traffic, finance, and weather~\citep{bu2020time,
bai2020adaptive, cheng2022financial, karevan2020transductive}.
Although modern forecasting models achieve strong in-distribution
performance~\citep{zeng2023transformers, nie2023a, liu2024itransformer},
their accuracy degrades after deployment as real-world time series
are non-stationary and the data-generating process drifts over
time~\citep{liu2022non, fan2023dish, ye2024frequency}.
Existing strategies such as online learning and domain adaptation either require labeled data, assume stable domain boundaries, or incur high computational costs, limiting their applicability to continuously evolving distributions~\citep{losing2018incremental,
wang2018deep}.
Complementary approaches based on normalization aim to address distribution shift through input transformation, but operate in a static manner and do not support feedback-driven adaptation over time~\citep{kim2021reversible,liu2023adaptive,liu2022non,fan2023dish,ye2024frequency}.

Test-time adaptation (TTA) offers a practical alternative by updating lightweight components of a frozen model using streaming test data~\citep{liang2025comprehensive}. Vision TTA methods~\citep{wang2021tent, sun2020test} rely on classification-specific assumptions and do not readily transfer to TSF. A key property shapes TTA design in TSF: ground-truth observations become available after the forecast horizon, enabling supervised adaptation by minimizing the forecasting loss. Recent TSF-TTA methods leverage this property to improve forecasting accuracy under distribution shift via lightweight online adaptation ~\citep{im2026cosa, grover2025shift,
kim2025battling, wang2026towards}.

\begin{figure}[t]
    \centering
    \begin{minipage}[b]{0.49\linewidth}
        \centering
        \includegraphics[width=\linewidth]{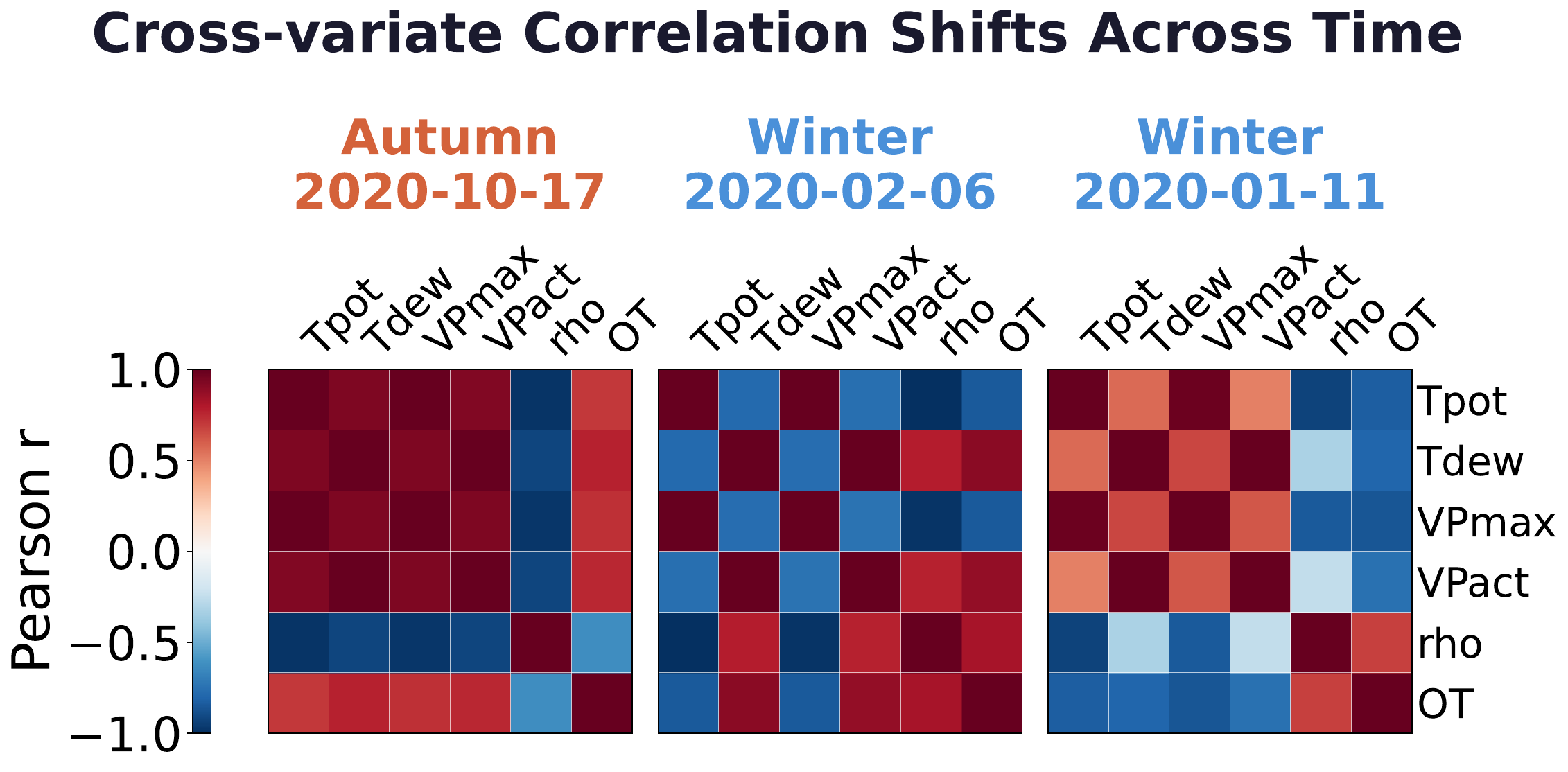}
        \captionof{figure}{Cross-variate dependencies in the Weather dataset.}
        \label{fig:motivation_a}
    \end{minipage}
    \hfill
    \begin{minipage}[b]{0.49\linewidth}
        \centering
        \includegraphics[width=\linewidth]{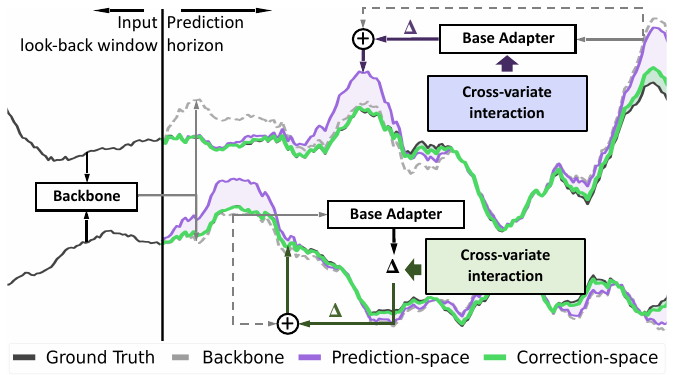}
        \captionof{figure}{Prediction-space vs.\ correction-space cross-variate interaction.}
        \label{fig:corrpred}
    \end{minipage}
\end{figure}

Existing TSF-TTA methods predominantly adapt each variate independently, yet distribution shifts in multivariate time series often affect variates jointly~\citep{lin2025cats} (Figure~\ref{fig:motivation_a}).
Such shifts may induce shared structure in the adaptation corrections required across variates, making cross-variate correction sharing potentially beneficial.
Our analysis shows that the leading singular mode captures approximately 73\% of correction energy on average (Appendix~\ref{app:rank_ablations}), further supporting this hypothesis.
Such shared structure motivates cross-variate interaction.
Prior work has explored \emph{how} to model this interaction. For example, \citet{im2026cosa} investigated
cross-variable attention and shared specific components on backbone predictions, yet found inconsistent benefits across datasets.
We identify a more fundamental design choice that has been overlooked: \emph{where} should cross-variate interaction take place?

\begin{figure}[t]
    \centering
    \includegraphics[width=0.94\linewidth]{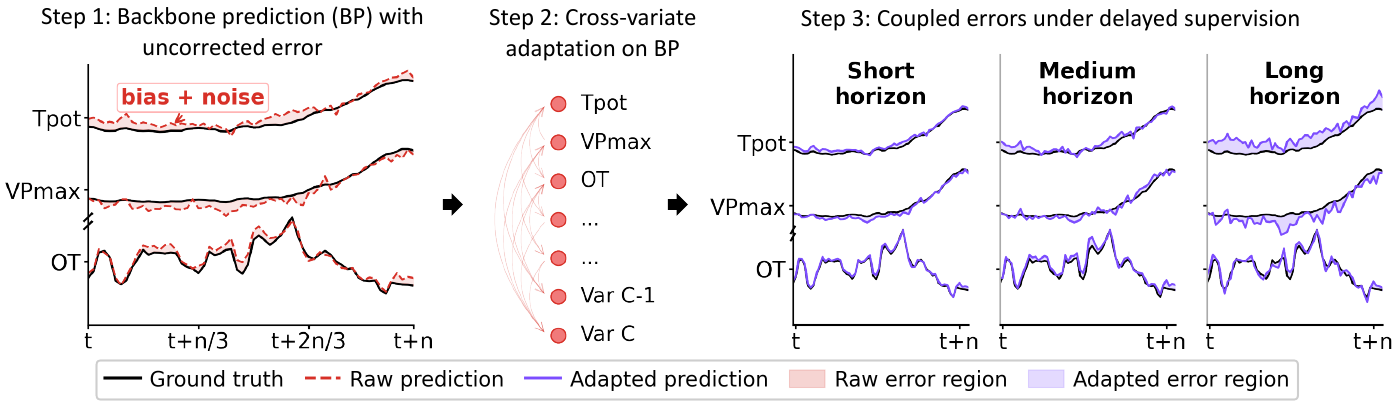}
    \caption{
    Direct error mixing in prediction-space cross-variate interaction. A conceptual illustration using variables from Weather (Tpot, VPmax, OT, and other variables). 
    Cross-variate interaction applied to backbone predictions can directly mix their uncorrected errors across variates (Step 2). In TSF-TTA, ground-truth observations become available only after the prediction horizon, delaying corrective feedback (Step 3). Proposition~\ref{prop:structure} formalizes the direct error-mixing pathway, while Figure~\ref{fig:main_horizon} empirically compares prediction-space and correction-space interaction across prediction horizons.
    }
    \label{fig:motivation_b}
\end{figure}

This choice of interaction space is particularly important in TSF-TTA under delayed supervision. Backbone predictions
made under distribution shift may contain substantial uncorrected errors.
Applying cross-variate interaction directly to these predictions introduces pathways through which errors from one variate can influence the refinement of others (Figure~\ref{fig:motivation_b}).
Since ground-truth observations become available only after the prediction horizon, such error propagation cannot be immediately corrected through supervised feedback.
We therefore investigate an alternative: performing cross-variate interaction on adapter-produced corrections (the \emph{correction space}) rather than on backbone predictions.
This design avoids directly mixing uncorrected backbone errors during cross-variate interaction.

In this work, we formalize the distinction between prediction- and correction-space interaction and characterize their different direct dependencies on frozen-backbone errors (Proposition~\ref{prop:structure}).
Building on this analysis, we propose \ours{} (\textsc{Co}rrection-space Interaction \textsc{Re}finement), a framework that performs structured cross-variate interaction in the correction space.
\ours{} introduces Shared-anchor Correction Refinement (SCR), which combines each variate's correction with a shared anchor through a parameter-efficient bottleneck. Input-conditioned spectral gating further modulates the refinement using information from the current input window.
Controlled experiments demonstrate the importance of interaction space, particularly at medium-to-long prediction horizons.
We summarize our contributions as follows:
\begin{enumerate}[nosep]
\item We identify the \emph{interaction space} as a key design choice for cross-variate TSF-TTA, showing that correction-space interaction enables information sharing across variates without directly mixing uncorrected backbone errors.

\item We propose \textsc{CoRe}, a framework to perform structured cross-variate adaptation in the \emph{correction space}, through \emph{Shared-anchor Correction Refinement (SCR)}, complemented by a lightweight spectral gate that modulates the refinement using the current input window.

\item Extensive experiments across seven backbones, six datasets, and four horizons demonstrate consistent improvements over existing TSF-TTA methods. Controlled comparisons show an advantage of correction-space over prediction-space interaction, and the gains are not explained by additional capacity alone.
\end{enumerate}

\section{Related Work}
\label{sec:related}
\textbf{Multivariate Time-Series Forecasting from a Channel Perspective.}
Multivariate time-series forecasting models temporal patterns in sequential data with multiple observed variables, whose interactions may vary across time and environments~\citep{qiu2025comprehensive}. 
There have been several approaches to modeling cross-variate relationships.
From a channel perspective, multivariate forecasters are often contrasted as channel-independent (CI)~\citep{nie2023a,zeng2023transformers,toner2024analysis} versus channel-dependent (CD)~\citep{zhou2021informer,wang2023micn,liu2024itransformer}, depending on whether they couple variables.
However, prior findings show that cross-variate modeling can be harmful when correlations are weak, noisy, or non-stationary, and may degrade robustness under distribution shift~\citep{chen2024calibration,he2025robust}.
These findings suggest that the question is not whether to model cross-variate dependencies, but how to introduce them without amplifying noise under distribution shift.

\textbf{Test-Time Adaptation for Time-Series Forecasting.}
Test-time adaptation is an emerging paradigm for addressing non-stationarity and distribution shifts in time-series forecasting.
In TSF-TTA, representative baselines include TAFAS~\citep{kim2025battling}, PETSA~\citep{medeiros2025accurate}, DynaTTA~\citep{grover2025shift}, and COSA~\citep{im2026cosa}. Recent work further introduces a matured-ground-truth protocol for TSF-TTA~\citep{wang2026towards}. Despite recent progress, existing methods typically adapt each variate independently, overlooking cross-variate structure.
Although cross-variate interaction on backbone predictions has been explored through attention and shared structures~\citep{im2026cosa}, 
its benefits are inconsistent across datasets, and operating on backbone predictions introduces a direct pathway for mixing their uncorrected errors across variates.
\ours{} instead performs cross-variate interaction in the \emph{correction space}, removing this direct backbone-error mixing pathway. See Appendix~\ref{app:extended_related} for further related work.

\section{Methodology}
\label{sec:method}

\subsection{Problem Formulation}
\label{sec:prelim}

We study test-time adaptation for multivariate time-series forecasting
under delayed supervision. At each time step $t$, a frozen backbone $f_\theta$ maps a past input window
$\mathbf{X}_t \in \mathbb{R}^{L \times C}$, with lookback length $L$ and
$C$ variates, to an $H$-horizon forecast
$\hat{\mathbf{Y}}_t^{\mathrm{base}} \in \mathbb{R}^{H \times C}$.
A lightweight adapter $\mathcal{A}_\phi$ produces per-variate corrections
$\boldsymbol{\Delta}_t
=
\mathcal{A}_\phi(\hat{\mathbf{Y}}_t^{\mathrm{base}},\mathbf{I}_t)
\in \mathbb{R}^{H \times C}$, where
$\boldsymbol{\Delta}_t^{(c)}\in\mathbb{R}^{H}$ denotes the correction for
variate $c$, and $\mathbf{I}_t$ denotes optional auxiliary information
available to the base adapter. The adapted forecast is
\begin{equation}
    \hat{\mathbf{Y}}_t
    =
    \hat{\mathbf{Y}}_t^{\mathrm{base}}
    +
    \boldsymbol{\Delta}_t.
    \label{eq:problem-definition}
\end{equation}
During deployment, ground-truth observations 
become available with a delay of
$H$ steps, and the adapter is updated only when the corresponding supervision becomes available. Thus, each prediction depends only on past observations, ensuring a leakage-free setting.
Adapter-specific instantiations and update protocols are
detailed in Appendix~\ref{app:baseline_config}
and~\ref{app:alternative_adapters}.

\vspace{-3pt}
\subsection{Cross-Variate Interaction in TSF-TTA}
\label{sec:correction_space}
Current TSF-TTA methods adapt each variate independently. Yet distribution shifts may induce shared structure in the corrections required across variates.
The decomposition in Equation~\ref{eq:problem-definition} offers two distinct places to model this cross-variate structure: the frozen-backbone prediction
$\hat{\mathbf{Y}}_t^{\mathrm{base}}$ and the adapter-produced correction
$\boldsymbol{\Delta}_t$.
Applying cross-variate interaction to each defines the \emph{prediction space} and the \emph{correction space} (Figure~\ref{fig:corrpred}). We analyze how backbone error propagates in each, and show that prediction-space interaction directly spreads this error across variates while correction-space interaction avoids such direct propagation.

Let $\mathcal{M}_{\mathrm{pred}}$ and $\mathcal{M}_{\mathrm{corr}}$
denote cross-variate refinement operators applied in the prediction and
correction spaces, respectively. 
The two interaction forms are

\begin{equation}
\begin{aligned}
\hat{\mathbf{Y}}_t^{\mathrm{pred}}
&=
\hat{\mathbf{Y}}_t^{\mathrm{base}}
+\boldsymbol{\Delta}_t
+\mathcal{M}_{\mathrm{pred}}(\hat{\mathbf{Y}}_t^{\mathrm{base}}),\\
\hat{\mathbf{Y}}_t^{\mathrm{corr}}
&=
\hat{\mathbf{Y}}_t^{\mathrm{base}}
+\boldsymbol{\Delta}_t
+\mathcal{M}_{\mathrm{corr}}(\boldsymbol{\Delta}_t).
\end{aligned}
\label{eq:interaction_spaces}
\end{equation}
Let $\mathbf{e}_t^{\mathrm{base}}
=\hat{\mathbf{Y}}_t^{\mathrm{base}}-\mathbf{Y}_t$ denote the frozen-backbone
error. Prediction-space interaction therefore operates on
$\mathbf{Y}_t+\mathbf{e}_t^{\mathrm{base}}$, whereas correction-space
interaction operates on $\boldsymbol{\Delta}_t$.

\begin{boxedproposition}[Structural Separation]
\label{prop:structure}
Define the backbone-error-induced prediction-space refinement as
\[
\boldsymbol{\Gamma}_t^{\mathrm{pred}}
:=
\mathcal{M}_{\mathrm{pred}}(\mathbf{Y}_t+\mathbf{e}_t^{\mathrm{base}})
-\mathcal{M}_{\mathrm{pred}}(\mathbf{Y}_t).
\]
For linear cross-variate mixing $\mathcal{M}_{\mathrm{pred}}(\cdot)=(\cdot)\mathbf{W}_{\mathrm{pred}}^\top$,
\[
\boldsymbol{\Gamma}_{t,:,c}^{\mathrm{pred}}
=
\sum_{j=1}^{C}
W_{\mathrm{pred},cj}\,\mathbf{e}_{t,:,j}^{\mathrm{base}}.
\]
Thus, each nonzero off-diagonal $W_{\mathrm{pred},cj}$ provides a direct
path from the backbone error of variate $j$ to the refinement of variate $c$.
The same structural distinction extends locally to nonlinear operators via
their Jacobians (Appendix~\ref{app:theory_formal}).
In correction space, the backbone error remains outside the cross-variate
operator.
\end{boxedproposition}

This proposition establishes a structural distinction in how the two interaction spaces directly depend on backbone errors. It does not imply that adapter corrections are independent of backbone errors, since the base adapter may itself depend on backbone predictions. Nor does it establish an accuracy ordering between the two interaction spaces. We investigate their empirical consequences in Section~\ref{sec:ablation_study}.

\subsection{The \ours{} Framework}
\label{sec:overview}

\begin{figure}[t]
    \centering
    \begin{overpic}[width=0.8\linewidth]{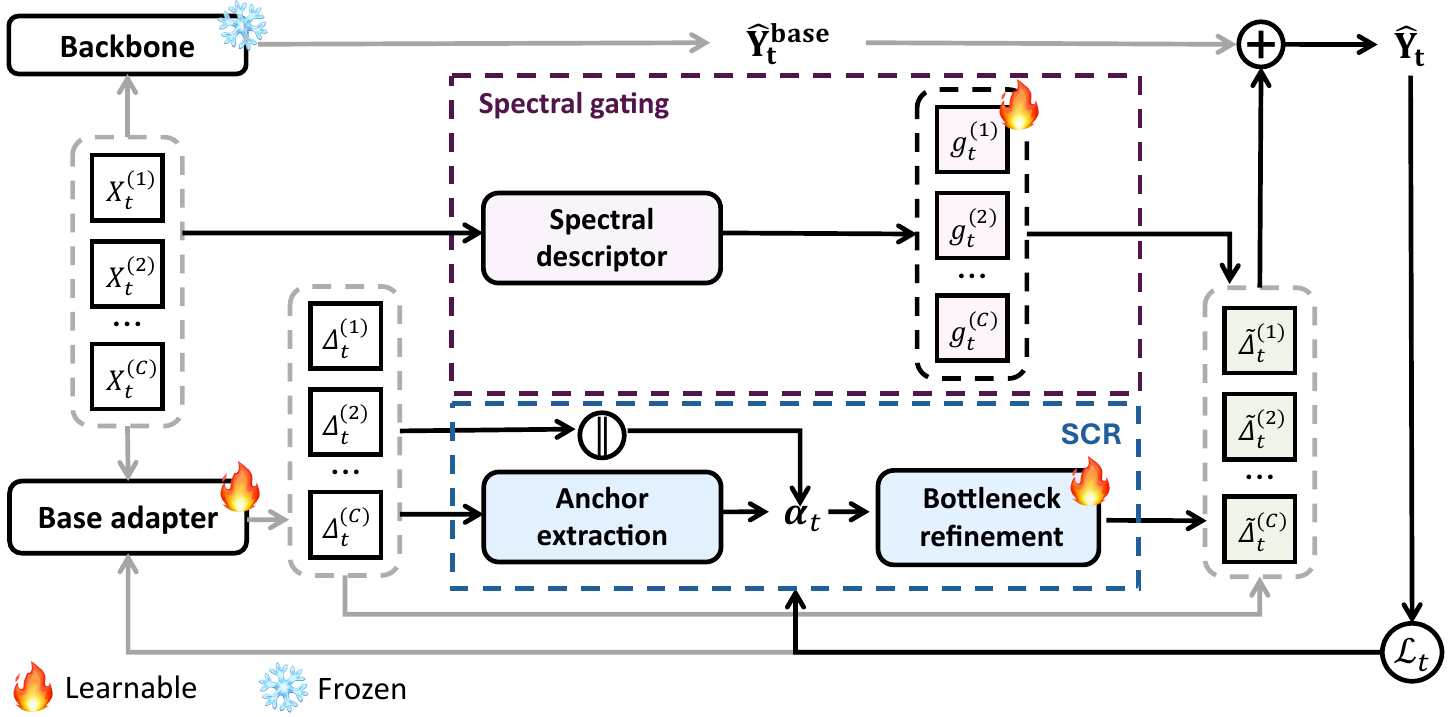}
        \put(15,84){\color{white}\rule{34mm}{4mm}}
        \put(15,84){\small Spectral gating}
    \end{overpic}
    \caption{Framework of \textsc{CoRe} for multivariate test-time adaptation. A frozen backbone produces a base forecast, while a base adapter produces per-variate corrections. SCR performs cross-variate interaction in the \emph{correction space}. Spectral descriptor of the current input window provides per-variate gates that modulate the SCR refinement. The refined corrections are added to the unchanged backbone prediction to form the final forecast.}
    \label{fig:design}
\end{figure}

Building on the above design principle, we propose \ours{} as an instantiation of cross-variate
adaptation in the correction space. The framework is illustrated in
Figure~\ref{fig:design}. Given the per-variate corrections produced by the
base adapter, \ours{} consists of two components: (i) \textbf{Shared-anchor Correction Refinement (SCR)} combines each
variate's correction with a shared correction anchor through a parameter-efficient
bottleneck; and (ii) \textbf{Input-Conditioned Spectral Gating} uses
spectral descriptors of the current input window to modulate the contribution of the SCR
refinement. The resulting refined corrections are added to the
backbone prediction to form the final forecast. The complete
leakage-free streaming procedure is given in
Appendix~\ref{app:implementation}.

\subsection{Shared-anchor Correction Refinement (SCR)}
\label{sec:scr}
SCR refines each variate's base-adapter correction by combining variate-specific and shared correction information. It consists of a shared correction anchor and a parameter-efficient bottleneck.

\noindent\textbf{Shared correction anchor.} Related distribution shifts may induce common patterns in the corrections required across variates. We approximate this shared information by averaging the corrections across variates:
\begin{equation}
    \boldsymbol{\alpha}_t
    =
    \frac{1}{C}
    \sum_{j=1}^{C}
    \boldsymbol{\Delta}_t^{(j)}
    \in\mathbb{R}^{H}.
    \label{eq:anchor}
\end{equation}
For each variate $c$, its own correction is concatenated with the shared
anchor:
\begin{equation}
    \mathbf{z}_t^{(c)}
    =
    [\boldsymbol{\Delta}_t^{(c)} \| \boldsymbol{\alpha}_t]
    \in\mathbb{R}^{2H},
    \label{eq:joint_rep}
\end{equation}
where $\|$ denotes concatenation. The concatenation preserves each variate's own correction while adding shared information.

\noindent\textbf{Bottleneck refinement.}
The joint representation is mapped to a cross-variate refinement through
a shared two-layer bottleneck:
\begin{equation}
    \boldsymbol{\delta}_t^{(c)}
    =
    \mathbf{W}_{\uparrow}
    \tanh\!\left(
        \mathbf{W}_{\downarrow}\mathbf{z}_t^{(c)}
    \right),
    \quad
    \mathbf{W}_{\downarrow}\in\mathbb{R}^{r\times2H},
    \quad
    \mathbf{W}_{\uparrow}\in\mathbb{R}^{H\times r}.
    \label{eq:bottleneck}
\end{equation}
where $\boldsymbol{\delta}_t^{(c)}\in\mathbb{R}^{H}$ is the refinement for
variate $c$, and $r$ is the bottleneck dimension.
Both projections $\mathbf{W}_{\downarrow}, \mathbf{W}_{\uparrow}$ are shared across variates.
We simply set $r=C$ by default,
giving $O(HC)$ SCR parameters against $O(H^2)$ for unconstrained dense mixing over the horizon. 
Sharing the anchor and constraining the bottleneck keep the interaction parameter-efficient, which mitigates overfitting under limited test-time supervision.

\subsection{Input-Conditioned Spectral Gating}
\label{sec:spectral}

SCR defines how corrections interact across variates.
Because temporal dynamics shift over time, the usefulness of shared correction information may also vary.
Spectral energy distribution provides useful cues for such changes~\citep{yi2023frequency, fan2024deep, piao2024frednormer} and can be computed directly from the current input window. We therefore use a lightweight spectral descriptor to modulate the SCR refinement:

\begin{equation}
    \widetilde{\boldsymbol{\Delta}}_t^{(c)}
    =
    \boldsymbol{\Delta}_t^{(c)}
    +
    g_t^{(c)}\boldsymbol{\delta}_t^{(c)},
    \label{eq:gated_refinement}
\end{equation}
where $\boldsymbol{\Delta}_t^{(c)}$ is the base-adapter correction,
$\boldsymbol{\delta}_t^{(c)}$ is the SCR refinement from
Equation~\ref{eq:bottleneck}, and $g_t^{(c)}\in(-1,1)$ provides signed modulation of the refinement for variate $c$.

\noindent\textbf{Spectral descriptor.}
For each variate in the input window $\mathbf{X}_t\in\mathbb{R}^{L\times C}$, we subtract its mean and compute a one-sided power spectrum via Fast Fourier Transform (FFT). We then 

average these across variates to obtain a shared spectral distribution
$\overline{\mathbf{P}}_t$ over $K_f=\lfloor L/2\rfloor+1$ frequency bins.

$\overline{\mathbf{P}}_t$ thus summarizes the window-level temporal dynamics across variates.
From $\overline{\mathbf{P}}_t$ we extract four spectral statistics.
One is the spectral entropy $\mathrm{SE}_t$, measuring how concentrated the spectral energy is~\citep{powell1979spectral,inouye1991quantification}. 
The other three are relative band powers over low-, mid-, and high-frequency bands $(\mathrm{LBR}_t,\mathrm{MBR}_t,\mathrm{HBR}_t)$, measuring how energy is distributed across frequency
ranges~\citep{keil2022recommendations}. Stacking these gives the spectral descriptor:

\begin{equation}
    \mathbf{s}_t
    =
    [\mathrm{SE}_t,\mathrm{LBR}_t,\mathrm{MBR}_t,\mathrm{HBR}_t]
    \in\mathbb{R}^{4}.
    \label{eq:spectral_descriptor}
\end{equation}
Normalization and frequency-band definitions are provided in
Appendix~\ref{app:spectral_desc}.

\noindent\textbf{Per-variate gating.}
We map the spectral descriptor to a $C$-dimensional gate:
\begin{equation}
    \mathbf{g}_t
    =
    \tanh\!\left(
        \mathbf{W}_g\mathbf{s}_t+\mathbf{b}_g
    \right)
    \in(-1,1)^C.
\label{eq:spectral_gate}
\end{equation}
Here $\mathbf{W}_g\in\mathbb{R}^{C\times4}$ and $\mathbf{b}_g\in\mathbb{R}^{C}$ are learnable.
The descriptor $\mathbf{s}_t$ is shared across variates, so the gates differ only through the rows of $\mathbf{W}_g$ and entries of $\mathbf{b}_g$.
The $\tanh$ bounds each coefficient to $(-1,1)$, allowing $g_t^{(c)}$ to provide signed modulation of the SCR refinement for variate $c$.

\subsection{Adaptation Objective}
\label{sec:objective}

The final adapted forecast is thus
$\hat{\mathbf{Y}}_t
=
\hat{\mathbf{Y}}_t^{\mathrm{base}}
+
\tilde{\boldsymbol{\Delta}}_t$.
The base-adapter parameters $\phi$ and CoRe parameters
$\Theta=\{\mathbf{W}_{\downarrow},\mathbf{W}_{\uparrow},\mathbf{W}_g,\mathbf{b}_g\}$
are updated jointly online.
At time $t$, let $\mathcal{B}_t$ denote the adaptation batch whose corresponding targets are available.
The parameters are updated by minimizing
\begin{equation}
\mathcal{L}_t
=
\frac{1}{|\mathcal{B}_t|}
\sum_{i\in\mathcal{B}_t}
\|\hat{\mathbf{Y}}_i-\mathbf{Y}_i\|_F^2
+
\lambda_{\mathrm{reg}}
\left(\|\phi\|_2^2+\|\Theta\|_2^2\right),
\label{eq:loss}
\end{equation}
where $\lambda_{\mathrm{reg}}$ controls weight decay to mitigate overfitting under limited supervision.

\section{Experiments}
\label{sec:experiments}

We evaluate \textsc{CoRe} from four perspectives: overall accuracy and efficiency, the choice of the interaction space, the design of correction-space refinement, and the reliability of online adaptation.

\subsection{Experimental Setup}
\label{sec:exp_setup}
\noindent\textbf{Datasets and evaluation.}
We evaluate on six multivariate forecasting benchmarks: ETTh1, ETTh2, ETTm1, ETTm2~\citep{zhou2021informer}, Weather,\footnote{\url{https://www.bgc-jena.mpg.de/wetter/}} and Exchange Rate~\citep{lai2018modeling}, with lookback $L{=}96$ and horizons $H\in\{96,192,336,720\}$. We additionally evaluate on high-dimensional datasets Electricity and Traffic~\citep{lai2018modeling} to assess scalability. We report MSE over the full test stream.

\noindent\textbf{Backbones and baselines.}
We use seven forecasting backbones: DLinear~\citep{zeng2023transformers}, FreTS~\citep{yi2023frequency}, OLS~\citep{toner2024analysis}, PatchTST~\citep{nie2023a}, iTransformer~\citep{liu2024itransformer}, Informer~\citep{zhou2021informer}, and MICN~\citep{wang2023micn}. 
We compare with the frozen backbone and TSF-TTA methods TAFAS~\citep{kim2025battling}, and COSA~\citep{im2026cosa}; PETSA~\citep{medeiros2025accurate} results are in Appendix~\ref{app:tsf_tta_methods}. 
All backbones use the same frozen checkpoints, and all TTA modules adapt under delayed supervision.
Unless stated otherwise, \textsc{CoRe} uses COSA as the base adapter, $r{=}C$, and the same hyperparameters across datasets and backbones. Experiments run on one NVIDIA A100; Appendix~\ref{app:implementation} gives implementation details.

\subsection{Overall Effectiveness}
\label{sec:main_results}
\begin{table*}[!ht]
\centering
\caption{Prediction MSE comparison (lower is better).
TAFAS, COSA and \ours{} results are averaged over 10 random seeds.
For \ours{}, standard deviations across seeds are small (maximum: 0.0130),
with full standard deviations reported in Table~\ref{tab:std}
(Appendix~\ref{app:seed_var}).
\textbf{Best}/\underline{second-best} highlighting applies to TAFAS vs. COSA vs.\ \ours{}.}
\label{tab:main_results_1}
{\normalsize
\setlength{\tabcolsep}{1.5pt}
\renewcommand{\arraystretch}{1.00}
\begin{adjustbox}{max width=\textwidth}
\begin{tabular}{cc cccc cccc cccc cccc}
\toprule
& & \multicolumn{4}{c}{\textbf{iTransformer}} 
& \multicolumn{4}{c}{\textbf{DLinear}} 
& \multicolumn{4}{c}{\textbf{FreTS}} 
& \multicolumn{4}{c}{\textbf{Informer}} \\
\cmidrule(lr){3-6} \cmidrule(lr){7-10} \cmidrule(lr){11-14} \cmidrule(lr){15-18}
& $H$ & Base & TAFAS & COSA & \ours{}
     & Base & TAFAS & COSA & \ours{}
     & Base & TAFAS & COSA & \ours{}
     & Base & TAFAS & COSA & \ours{} \\
\midrule
\multirow{4}{*}{\rotatebox[origin=c]{90}{\textbf{ETTh1}}}
& 96  
& 0.4580 & \textbf{0.4381} & 0.4638 & \underline{0.4545}
& 0.4695 & \textbf{0.4604} & 0.4922 & \underline{0.4916}
& 0.4462 & \textbf{0.4395} & \underline{0.4623} & 0.4625
& 0.7466 & 0.6879 & \underline{0.6589} & \textbf{0.6501} \\
& 192 
& 0.5070 & \underline{0.4911} & 0.5024 & \textbf{0.4589}
& 0.5213 & \underline{0.5099} & 0.5371 & \textbf{0.4961}
& 0.5021 & \underline{0.4944} & 0.5084 & \textbf{0.4692}
& 0.9127 & 0.7905 & \underline{0.6790} & \textbf{0.6415} \\
& 336 
& 0.5788 & 0.5696 & \underline{0.4956} & \textbf{0.4661}
& 0.5659 & 0.5622 & \underline{0.5208} & \textbf{0.4722}
& 0.5544 & 0.5560 & \underline{0.4970} & \textbf{0.4596}
& 1.1265 & 0.9582 & \underline{0.6727} & \textbf{0.6302} \\
& 720 
& 0.7067 & 0.6687 & \underline{0.5381} & \textbf{0.5055}
& 0.6992 & 0.6685 & \underline{0.5433} & \textbf{0.4891}
& 0.7145 & 0.6838 & \underline{0.5616} & \textbf{0.5052}
& 1.3596 & 0.9986 & \underline{0.7123} & \textbf{0.6400} \\
\midrule
\multirow{4}{*}{\rotatebox[origin=c]{90}{\textbf{ETTh2}}}
& 96  
& 0.2731 & \textbf{0.2701} & 0.3050 & \underline{0.2803}
& 0.2323 & \textbf{0.2302} & 0.2552 & \underline{0.2426}
& 0.2386 & \textbf{0.2362} & 0.2560 & \underline{0.2462}
& 0.3510 & \textbf{0.3419} & \underline{0.3747} & 0.3760 \\
& 192 
& 0.3088 & \underline{0.3042} & 0.3116 & \textbf{0.2722}
& 0.2862 & 0.2836 & \underline{0.2739} & \textbf{0.2275}
& 0.2867 & 0.2829 & \underline{0.2676} & \textbf{0.2371}
& 0.4390 & 0.4311 & \underline{0.4197} & \textbf{0.3567} \\
& 336 
& 0.3567 & 0.3397 & \underline{0.2552} & \textbf{0.2245}
& 0.3252 & 0.3193 & \underline{0.2521} & \textbf{0.2171}
& 0.3317 & 0.3221 & \underline{0.2548} & \textbf{0.2196}
& 0.4884 & 0.4900 & \underline{0.4127} & \textbf{0.3463} \\
& 720 
& 0.4297 & 0.4128 & \underline{0.2695} & \textbf{0.2545}
& 0.4149 & 0.3939 & \underline{0.2589} & \textbf{0.2324}
& 0.4198 & 0.3976 & \underline{0.2516} & \textbf{0.2336}
& 0.5206 & 0.5172 & \underline{0.3297} & \textbf{0.3113} \\
\midrule
\multirow{4}{*}{\rotatebox[origin=c]{90}{\textbf{ETTm1}}}
& 96  
& 0.3740 & \textbf{0.3538} & 0.3799 & \underline{0.3776}
& 0.3715 & \textbf{0.3488} & \underline{0.3800} & 0.3881
& 0.3676 & \textbf{0.3575} & \underline{0.3836} & 0.3862
& 0.6475 & 0.4987 & \underline{0.4956} & \textbf{0.4929} \\
& 192 
& 0.4421 & \textbf{0.4175} & 0.4346 & \underline{0.4184}
& 0.4438 & \textbf{0.4159} & 0.4755 & \underline{0.4282}
& 0.4325 & \underline{0.4206} & 0.4615 & \textbf{0.4200}
& 0.6950 & \underline{0.5472} & 0.5571 & \textbf{0.5385} \\
& 336 
& 0.5096 & \underline{0.4727} & 0.4825 & \textbf{0.4585}
& 0.5183 & \underline{0.4787} & 0.5358 & \textbf{0.4665}
& 0.5006 & \underline{0.4816} & 0.5195 & \textbf{0.4616}
& 0.7708 & 0.6517 & \underline{0.6031} & \textbf{0.5771} \\
& 720 
& 0.6083 & \underline{0.5539} & 0.5615 & \textbf{0.5171}
& 0.5920 & \underline{0.5495} & 0.5920 & \textbf{0.5424}
& 0.5590 & \underline{0.5413} & 0.5773 & \textbf{0.5200}
& 0.8308 & 0.7888 & \underline{0.6713} & \textbf{0.6454} \\
\midrule
\multirow{4}{*}{\rotatebox[origin=c]{90}{\textbf{ETTm2}}}
& 96  
& 0.1791 & \textbf{0.1722} & 0.2226 & \underline{0.1831}
& 0.1598 & \textbf{0.1588} & 0.1915 & \underline{0.1644}
& 0.1582 & \textbf{0.1573} & 0.1900 & \underline{0.1707}
& 0.2839 & \underline{0.2176} & 0.2253 & \textbf{0.2027} \\
& 192 
& 0.2171 & \underline{0.2153} & 0.2393 & \textbf{0.1905}
& 0.1930 & \underline{0.1927} & 0.1961 & \textbf{0.1652}
& 0.1922 & 0.1911 & \underline{0.1885} & \textbf{0.1628}
& 0.4291 & 0.2732 & \underline{0.2408} & \textbf{0.1964} \\
& 336 
& 0.2722 & 0.2644 & \underline{0.2473} & \textbf{0.2058}
& 0.2324 & 0.2337 & \underline{0.1969} & \textbf{0.1725}
& 0.2319 & 0.2308 & \underline{0.1944} & \textbf{0.1686}
& 0.4123 & 0.3676 & \underline{0.2660} & \textbf{0.2114} \\
& 720 
& 0.3321 & 0.3216 & \underline{0.2329} & \textbf{0.1955}
& 0.3062 & 0.3053 & \underline{0.2066} & \textbf{0.1778}
& 0.3010 & 0.2984 & \underline{0.2110} & \textbf{0.1799}
& 0.5985 & 0.4416 & \underline{0.2804} & \textbf{0.2307} \\
\midrule
\multirow{4}{*}[0.4ex]{\rotatebox[origin=c]{90}{\textbf{\shortstack{Exchange\\Rate}}}}
& 96  
& 0.0918 & 0.0918 & \underline{0.0913} & \textbf{0.0832}
& 0.0913 & 0.0878 & \underline{0.0816} & \textbf{0.0706}
& 0.0828 & \underline{0.0790} & 0.0835 & \textbf{0.0738}
& 0.1720 & 0.1605 & \underline{0.1187} & \textbf{0.0985} \\
& 192 
& 0.2086 & 0.1997 & \underline{0.1093} & \textbf{0.0988}
& 0.1827 & 0.1736 & \underline{0.0844} & \textbf{0.0733}
& 0.1733 & 0.1637 & \underline{0.0959} & \textbf{0.0806}
& 0.3080 & 0.2783 & \underline{0.1265} & \textbf{0.1183} \\
& 336 
& 0.3414 & 0.3119 & \underline{0.1192} & \textbf{0.1054}
& 0.3277 & 0.3067 & \underline{0.0867} & \textbf{0.0804}
& 0.3241 & 0.3005 & \underline{0.0978} & \textbf{0.0933}
& 0.6753 & 0.5693 & \underline{0.1902} & \textbf{0.1901} \\
& 720 
& 0.9137 & 0.9062 & \underline{0.2228} & \textbf{0.1957}
& 0.8383 & 0.8281 & \underline{0.1469} & \textbf{0.1239}
& 0.8267 & 0.8142 & \underline{0.1537} & \textbf{0.1289}
& 1.5502 & 1.5732 & \underline{0.3524} & \textbf{0.3403} \\
\midrule
\multirow{4}{*}{\rotatebox[origin=c]{90}{\textbf{Weather}}}
& 96  
& 0.1708 & \textbf{0.1625} & 0.1753 & \underline{0.1635}
& 0.1954 & 0.1828 & \underline{0.1814} & \textbf{0.1719}
& 0.1857 & 0.1743 & \underline{0.1739} & \textbf{0.1617}
& 0.2168 & \textbf{0.1762} & 0.1940 & \underline{0.1807} \\
& 192 
& 0.2208 & 0.2086 & \underline{0.2000} & \textbf{0.1697}
& 0.2403 & 0.2219 & \underline{0.1970} & \textbf{0.1711}
& 0.2311 & 0.2143 & \underline{0.1944} & \textbf{0.1690}
& 0.3344 & \underline{0.2567} & 0.2627 & \textbf{0.2261} \\
& 336 
& 0.2807 & 0.2619 & \underline{0.2439} & \textbf{0.1736}
& 0.2918 & 0.2695 & \underline{0.2465} & \textbf{0.1782}
& 0.2844 & 0.2639 & \underline{0.2341} & \textbf{0.1785}
& 0.3793 & 0.3076 & \underline{0.2773} & \textbf{0.2202} \\
& 720 
& 0.3572 & 0.3510 & \underline{0.2193} & \textbf{0.1876}
& 0.3643 & 0.3550 & \underline{0.2231} & \textbf{0.1777}
& 0.3555 & 0.3509 & \underline{0.2084} & \textbf{0.1763}
& 0.5370 & 0.4057 & \underline{0.2674} & \textbf{0.2371} \\
\bottomrule
\end{tabular}
\end{adjustbox}
}
\end{table*}
\begin{table*}[!ht]
\centering
\vspace{-10pt}% tighten the visual gap from the previous part of Table 1
\ContinuedFloat
% (No repeated caption here; keeps the continuation tight)
{\scriptsize
\setlength{\tabcolsep}{1.5pt}
\renewcommand{\arraystretch}{1.00}
\begin{adjustbox}{max width=\textwidth,scale=0.9}
\begin{tabular}{cc cccc cccc cccc}
\toprule
& & \multicolumn{4}{c}{\textbf{PatchTST}} 
& \multicolumn{4}{c}{\textbf{OLS}} 
& \multicolumn{4}{c}{\textbf{MICN}} \\
\cmidrule(lr){3-6} \cmidrule(lr){7-10} \cmidrule(lr){11-14}
& $H$ & Base & TAFAS & COSA & \ours{}
     & Base & TAFAS & COSA & \ours{}
     & Base & TAFAS & COSA & \ours{} \\
\midrule
\multirow{4}{*}{\rotatebox[origin=c]{90}{\textbf{ETTh1}}}
& 96  
& 0.4329 & \textbf{0.4264} & 0.4392 & \underline{0.4371}
& 0.4511 & \textbf{0.4394} & 0.4729 & \underline{0.4723}
& 0.6244 & \textbf{0.5836} & \underline{0.5837} & 0.5863 \\
& 192 
& 0.4910 & 0.4820 & \underline{0.4738} & \textbf{0.4540}
& 0.5046 & \underline{0.4918} & 0.5173 & \textbf{0.4763}
& 0.6543 & 0.5902 & \underline{0.5802} & \textbf{0.5203} \\
& 336 
& 0.5554 & 0.5463 & \underline{0.4971} & \textbf{0.4458}
& 0.5510 & 0.5453 & \underline{0.5041} & \textbf{0.4568}
& 0.7352 & 0.6767 & \underline{0.5889} & \textbf{0.5386} \\
& 720 
& 0.7009 & 0.6826 & \underline{0.5449} & \textbf{0.4983}
& 0.6996 & 0.6619 & \underline{0.5443} & \textbf{0.4886}
& 0.9336 & 0.7926 & \underline{0.6177} & \textbf{0.5591} \\
\midrule
\multirow{4}{*}{\rotatebox[origin=c]{90}{\textbf{ETTh2}}}
& 96  
& 0.2355 & \textbf{0.2349} & 0.2506 & \underline{0.2490}
& 0.2305 & \textbf{0.2290} & 0.2443 & \underline{0.2389}
& 0.2537 & \underline{0.2498} & 0.2538 & \textbf{0.2476} \\
& 192 
& 0.2855 & \underline{0.2765} & 0.2767 & \textbf{0.2405}
& 0.2838 & 0.2824 & \underline{0.2740} & \textbf{0.2290}
& 0.3082 & 0.3077 & \underline{0.2923} & \textbf{0.2430} \\
& 336 
& 0.3174 & 0.3108 & \underline{0.2451} & \textbf{0.2167}
& 0.3257 & 0.3202 & \underline{0.2448} & \textbf{0.2122}
& 0.3738 & 0.3653 & \underline{0.2828} & \textbf{0.2486} \\
& 720 
& 0.4063 & 0.4039 & \underline{0.2514} & \textbf{0.2331}
& 0.4165 & 0.3978 & \underline{0.2499} & \textbf{0.2301}
& 0.4611 & 0.4591 & \underline{0.3058} & \textbf{0.2746} \\
\midrule
\multirow{4}{*}{\rotatebox[origin=c]{90}{\textbf{ETTm1}}}
& 96  
& 0.4032 & \textbf{0.3908} & 0.4057 & \underline{0.4026}
& 0.3706 & \textbf{0.3495} & \underline{0.3780} & 0.3874
& 0.4352 & \textbf{0.3975} & 0.4240 & \underline{0.4232} \\
& 192 
& 0.4508 & \underline{0.4363} & 0.4417 & \textbf{0.4183}
& 0.4438 & \textbf{0.4148} & 0.4730 & \underline{0.4235}
& 0.4940 & \underline{0.4587} & 0.4822 & \textbf{0.4551} \\
& 336 
& 0.5040 & \underline{0.4855} & 0.4989 & \textbf{0.4509}
& 0.5177 & \underline{0.4769} & 0.5360 & \textbf{0.4658}
& 0.5605 & \underline{0.5136} & 0.5480 & \textbf{0.5018} \\
& 720 
& 0.5607 & \underline{0.5441} & 0.5696 & \textbf{0.5225}
& 0.5927 & \underline{0.5481} & 0.5938 & \textbf{0.5367}
& 0.6169 & \underline{0.5726} & 0.6061 & \textbf{0.5602} \\
\midrule
\multirow{4}{*}{\rotatebox[origin=c]{90}{\textbf{ETTm2}}}
& 96  
& 0.1593 & \textbf{0.1598} & 0.1953 & \underline{0.1721}
& 0.1602 & \textbf{0.1596} & 0.1923 & \underline{0.1655}
& 0.1697 & \textbf{0.1724} & 0.1909 & \underline{0.1845} \\
& 192 
& 0.2066 & 0.2052 & \underline{0.2023} & \textbf{0.1694}
& 0.1935 & \underline{0.1940} & 0.1978 & \textbf{0.1671}
& 0.2198 & \underline{0.2125} & 0.2307 & \textbf{0.1832} \\
& 336 
& 0.2480 & 0.2530 & \underline{0.2100} & \textbf{0.1765}
& 0.2332 & 0.2348 & \underline{0.1985} & \textbf{0.1726}
& 0.2603 & 0.2590 & \underline{0.2238} & \textbf{0.1871} \\
& 720 
& 0.3057 & 0.3045 & \underline{0.2058} & \textbf{0.1728}
& 0.3066 & 0.3069 & \underline{0.2067} & \textbf{0.1787}
& 0.3352 & 0.3305 & \underline{0.2381} & \textbf{0.1991} \\
\midrule
\multirow{4}{*}[1.1ex]{\rotatebox[origin=c]{90}{\textbf{\shortstack{Exchange\\Rate}}}}
& 96  
& 0.0858 & 0.0817 & \underline{0.0798} & \textbf{0.0739}
& 0.0814 & \underline{0.0789} & 0.0795 & \textbf{0.0741}
& 0.1114 & 0.1100 & \underline{0.0920} & \textbf{0.0779} \\
& 192 
& 0.1880 & 0.1760 & \underline{0.1029} & \textbf{0.0897}
& 0.1727 & 0.1643 & \underline{0.0953} & \textbf{0.0812}
& 0.2081 & 0.1864 & \underline{0.0855} & \textbf{0.0739} \\
& 336 
& 0.3383 & 0.3088 & \underline{0.1094} & \textbf{0.1000}
& 0.3225 & 0.3014 & \underline{0.1003} & \textbf{0.0947}
& 0.3941 & 0.3644 & \underline{0.1299} & \textbf{0.1091} \\
& 720 
& 0.8327 & 0.8330 & \underline{0.1743} & \textbf{0.1348}
& 0.8356 & 0.8202 & \underline{0.1490} & \textbf{0.1259}
& 1.4477 & 0.8503 & \underline{0.2320} & \textbf{0.2211} \\
\midrule
\multirow{4}{*}{\rotatebox[origin=c]{90}{\textbf{Weather}}}
& 96  
& 0.1727 & 0.1702 & \underline{0.1664} & \textbf{0.1622}
& 0.1958 & 0.1831 & \underline{0.1812} & \textbf{0.1704}
& 0.1724 & 0.1723 & \underline{0.1701} & \textbf{0.1646} \\
& 192 
& 0.2197 & 0.2145 & \underline{0.1939} & \textbf{0.1724}
& 0.2405 & 0.2220 & \underline{0.1960} & \textbf{0.1719}
& 0.2235 & 0.2215 & \underline{0.2078} & \textbf{0.1608} \\
& 336 
& 0.2763 & \underline{0.2655} & 0.2834 & \textbf{0.1810}
& 0.2922 & 0.2699 & \underline{0.2440} & \textbf{0.1777}
& 0.2815 & 0.2775 & \underline{0.2107} & \textbf{0.1698} \\
& 720 
& 0.3517 & 0.3350 & \underline{0.2056} & \textbf{0.1765}
& 0.3645 & 0.3550 & \underline{0.2252} & \textbf{0.1785}
& 0.3507 & 0.3682 & \underline{0.2013} & \textbf{0.1711} \\
\bottomrule
\end{tabular}
\end{adjustbox}
}
\end{table*}

Table~\ref{tab:main_results_1} reports MSE across all 168 settings ($7 \text{ backbones}\times6 \text{ datasets} \times4\text{ horizons}$). Averaged across all settings, \textsc{CoRe} reduces MSE by $25.82\%$ relative to the frozen backbone and by $10.57\%$ relative to COSA. The largest dataset-level reductions over COSA occur on Weather, ETTm2, and Exchange Rate.
Gains are more pronounced at medium-to-long horizons: $5.27\%$ at $H{=}96$, $13.34\%$ at $H{=}336$, and $11.53\%$ at $H{=}720$. The few underperforming cases occur mainly at $H{=}96$, where the benefit of correction-space interaction may be smaller.
\textsc{CoRe} applies larger corrections where the frozen backbone is less accurate ($167/168$ settings) and yields larger relative improvements there ($166/168$); see Appendix~\ref{app:backbone_quality}.

\noindent\textbf{Runtime overhead.}
Averaged over six datasets and four horizons with DLinear, \textsc{CoRe}
increases the per-adaptation-step runtime from $6.28$\,ms to $6.76$\,ms
over COSA ($+0.48$\,ms; $+7.7\%$). Thus, the accuracy gains come with
modest additional computation; Appendix~\ref{app:timing} provides the
full timing breakdown.

\subsection{Why Correction-Space Interaction?}
\label{sec:ablation_study}

\noindent\textbf{Correction vs. prediction space.}
Correction-space interaction shows a clear advantage over prediction-space interaction at medium-to-long horizons. Figure~\ref{fig:main_horizon} controls for the interaction architecture, changing only the space in which cross-variate interaction is performed. While the two variants are close at $H{=}96$, at $H{=}336$ correction-space interaction improves DLinear by $33.3\%$, compared with $19.8\%$ for prediction-space interaction. The same overall horizon-dependent pattern holds across all seven backbones (Appendix~\ref{app:corrvspred}). Together with Proposition~\ref{prop:structure}, this controlled comparison supports the empirical advantage of correction-space over prediction-space interaction, particularly at medium-to-long horizons.

\noindent\textbf{Parameter-matched comparison.}
The advantage of \textsc{CoRe} is not explained by additional trainable capacity alone. Table~\ref{tab:param_matched} compares \textsc{CoRe} with a parameter-matched independent COSA+MLP, which adds per-variate capacity without cross-variate interaction. COSA+MLP improves over COSA by only $0.97\%$ on average (single seed; Table~\ref{tab:param_matched}), compared with $10.55\%$ for \textsc{CoRe}. The contrast is particularly clear on Exchange Rate, where cross-variate relationships vary substantially over time (Appendix~\ref{app:dataset_correlation}). These results support the benefit of structured cross-variate refinement under a comparable parameter budget.

\begin{table*}[t]
\begin{minipage}[t]{0.49\textwidth}
  \centering
  \captionsetup[table]{width=\linewidth,singlelinecheck=false}
  \small
  \setlength{\tabcolsep}{2pt}
  \renewcommand{\arraystretch}{0.8}
  \captionof{table}{Parameter-matched comparison (DLinear, 6 datasets $\times$ 4 horizons; seed = 42).
  Exchange Rate reported separately due to its high non-stationarity.}
  \label{tab:param_matched}
  \begin{tabular*}{\linewidth}{@{\extracolsep{\fill}}lrcc}
  \toprule
  \textbf{Method} & \textbf{Params} & \textbf{Avg. MSE} & \textbf{Exch. MSE} \\
  \midrule
  COSA & 1,627,930 & 0.2974 & 0.0983 \\
  COSA+MLP & 1,637,575 & 0.2945 & 0.0986 \\
  \midrule
  \textsc{CoRe} & 1,637,898 & 0.2660 & 0.0849 \\
  \bottomrule
  \end{tabular*}
\end{minipage}
\hfill
\begin{minipage}[t]{0.49\textwidth}
  \centering
  \captionsetup[table]{width=\linewidth,singlelinecheck=false}
  
  \setlength{\tabcolsep}{2pt}
  \renewcommand{\arraystretch}{0.8}
  \captionof{table}{Results on large-scale datasets Electricity and Traffic, averaged over 4 horizons and 10 seeds. All methods use a fixed bottleneck rank $r{=}16$. }
  \label{tab:highdimshort}
  \resizebox{0.85\linewidth}{!}{%
  \begin{tabular}{llccc}
  \toprule
  \textbf{Model} & \textbf{Dataset} &
  \textbf{Backbone} & \textbf{COSA} &
  \textbf{\textsc{CoRe}} \\
  \midrule
  \multirow{2}{*}{DLinear}
  & Electricity & 0.2258 & 0.2008 & \textbf{0.1860} \\
  & Traffic     & 0.6502 & 0.6352 & \textbf{0.6325} \\
  \midrule
  \multirow{2}{*}{MICN}
  & Electricity & 0.1962 & 0.1619 & \textbf{0.1458} \\
  & Traffic     & 0.5275 & 0.4875 & \textbf{0.4694} \\
  \bottomrule
  \end{tabular}%
  }
\end{minipage}
\end{table*}

\subsection{Structured and Selective Refinement}
\label{sec:design_components}
\begin{table}[!ht] 
\centering
\caption{Ablation study of \textsc{CoRe} components.
Average MSE over three backbones (DLinear, PatchTST, MICN),
six datasets, four horizons, and 10 seeds.
\textbf{vs.\ COSA}: relative MSE reduction over COSA.
\emph{Loss-trend gate}: COSA's adaptive learning-rate
signal repurposed as the SCR gate.
\emph{Fixed gate} ($g{=}1$): uniform cross-variate
refinement without input-dependent modulation.
$\mathbf{s}_t$: Spectral descriptors for gating (Section~\ref{sec:spectral}).}
\label{tab:ablation_main}
\setlength{\tabcolsep}{2pt}
\renewcommand{\arraystretch}{0.85}
\resizebox{\linewidth}{!}{%
\begin{tabular}{lccccc}
\toprule
\textbf{Method} & 
\textbf{DLinear} & \textbf{PatchTST} & \textbf{MICN} &
\textbf{Avg.\ MSE} & \textbf{vs.\ COSA} \\
\midrule
COSA & 0.2981 & 0.2927 & 0.3241 & 0.3050 & -- \\
\midrule
+SCR (no anchor; per-variate bottleneck) &
0.2760 & 0.2721 & 0.3018 & 0.2833 & $-7.1\%$ \\
+SCR (no anchor/bottleneck; full $C{\times}C$ mixing) &
0.2745 & 0.2706 & 0.3044 & 0.2832 & $-7.2\%$ \\
\midrule
+SCR (loss-trend gate) & 
0.2712 & 0.2657 & 0.2963 & 0.2777 & $-9.0\%$ \\
+SCR (fixed gate, $g{=}1$) & 
0.2703 & 0.2651 & 0.2954 & 0.2769 & $-9.2\%$ \\
\midrule

\textsc{CoRe} & 
0.2675 & 0.2646 & 0.2942 & 
0.2754 & $-9.7\%$ \\
\bottomrule
\end{tabular}%
}
\end{table}

\noindent\textbf{Shared and low-dimensional correction structure.}
As shown in Table~\ref{tab:ablation_main}, the full SCR design improves over COSA by $9.7\%$, compared with $7.1\%$ without the shared anchor and $7.2\%$ with unconstrained $C\times C$ mixing. These results support the benefit of the shared-anchor structure over the evaluated alternatives. 

Moreover, the leading singular mode captures $73\%$ of the correction energy on average, revealing pronounced low-dimensional structure. 
With $r{=}16\ll C$, SCR retains improvements over COSA on both Electricity ($C{=}321$) and Traffic ($C{=}862$), showing that its effectiveness does not require the bottleneck rank to scale with the number of variates (Table~\ref{tab:highdimshort}). Together, these results support the shared-anchor and parameter-efficient bottleneck designs in SCR. See  Appendix~\ref{app:rank_ablations} for the rank and scalability analysis.

\noindent\textbf{Spectral gating.} 
The spectral signal is computed directly from the current input window and is therefore available from the first test batch, without requiring the historical loss or model-state statistics used by loss-trend or embedding-drift signals.
Table~\ref{tab:ablation_main} shows that with spectral gating, \textsc{CoRe} improves over COSA by $9.7\%$, compared with $9.2\%$ for a fixed gate and $9.0\%$ for loss-trend gating. Spectral gating thus provides a modest additional gain, while most of the improvement comes from SCR. Appendix~\ref{app:spectral_full_ablations} contains the full ablations.

\subsection{Adaptation Reliability}
\label{sec:reliability}

\begin{figure}[t]
    \centering
    \begin{minipage}[t]{0.58\linewidth}
        \vspace{0pt}
        \centering
        \includegraphics[width=\linewidth]{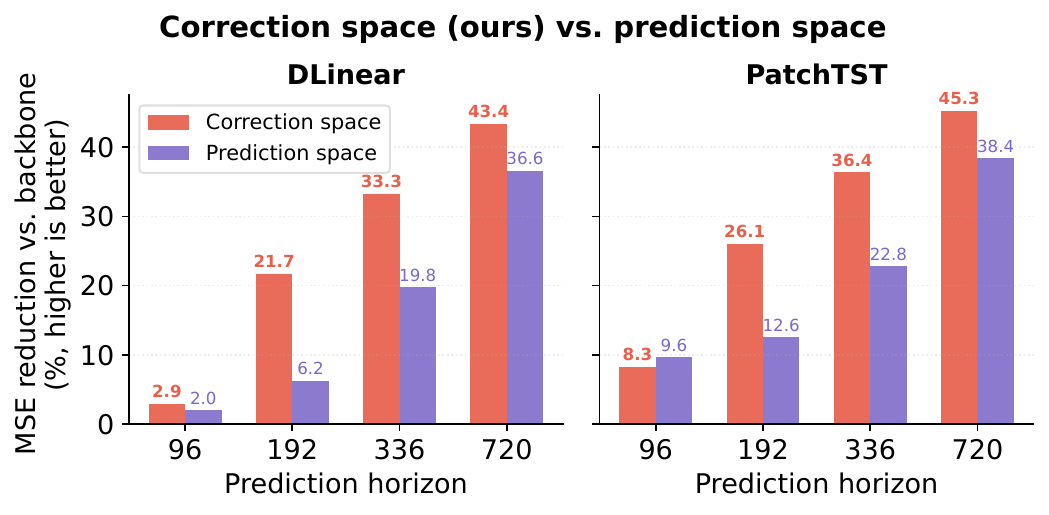}
        \captionof{figure}{Relative MSE reduction over the frozen backbone for
        correction-space and prediction-space interaction under the same
        bottleneck architecture.}
        \label{fig:main_horizon}
    \end{minipage}
    \hfill
    \begin{minipage}[t]{0.40\linewidth}
        \vspace{0pt}
        \centering
        \footnotesize
        \setlength{\tabcolsep}{3.2pt}
        \renewcommand{\arraystretch}{1.12}
        \captionof{table}{Online adaptation reliability over 18 backbone-dataset combinations (3 backbones $\times$ 6 datasets), each averaged over $H\in\{96,192,336,720\}$. Lower is better; darker blue indicates better performance within each column.}
        \label{tab:shortregret}
        \vspace{2pt}
        \begin{tabular}{lcccc}
        \toprule
        \textbf{Method}
        & $\overline{\mathcal{R}}_T\!\downarrow$
        & $p_{\mathrm{neg}}\!\downarrow$
        & $\overline{\mathcal{R}}_T^{+}\!\downarrow$
        & $n_{\mathrm{worse}}\!\downarrow$ \\
        \midrule
        TAFAS
        & $-0.010$
        & $0.446$
        & $0.0138$
        & $8$ \\

        COSA
        & \cellcolor{blue!8}$-0.087$
        & \cellcolor{blue!8}$0.277$
        & \cellcolor{blue!8}$0.0127$
        & \cellcolor{blue!8}$5$ \\

        \textsc{CoRe}
        & \cellcolor{blue!20}$\mathbf{-0.117}$
        & \cellcolor{blue!20}$\mathbf{0.232}$
        & \cellcolor{blue!20}$\mathbf{0.0108}$
        & \cellcolor{blue!20}$\mathbf{2}$ \\
        \bottomrule
        \end{tabular}
    \end{minipage}
\end{figure}

Online updates aim to improve performance but can also degrade it~\citep{niu2023towards}. Reliable adaptation therefore matters alongside average accuracy, particularly under delayed supervision where corrective feedback is not immediately available.
\textsc{CoRe}'s accuracy gains are accompanied by more reliable online adaptation. 
We assess reliability on three representative backbones (DLinear, PatchTST, and MICN) using four complementary measures:
average correction regret ($\overline{\mathcal R}_T$), the frequency of
harmful adaptation ($p_{\mathrm{neg}}$), the severity of harmful adaptation
($\overline{\mathcal R}_T^{+}$), and the number of settings in which
adaptation underperforms the frozen backbone ($n_{\mathrm{worse}}$).
Table~\ref{tab:shortregret} shows that \textsc{CoRe} improves over COSA across all four reliability measures.
Together, these results show that \textsc{CoRe} makes online adaptation more consistently beneficial relative to the frozen backbone.
Appendix~\ref{app:eval_framework} provides the metric definitions and
detailed breakdowns.

\section{Conclusion}

We presented \textsc{CoRe}, a multivariate TSF-TTA framework centered on cross-variate interaction in the \emph{correction space}.
By performing interaction on adapter-produced corrections rather than backbone predictions, 
\textsc{CoRe} avoids 
directly mixing uncorrected backbone errors during cross-variate interaction. The principle is instantiated through structured Shared-anchor Correction Refinement, complemented by lightweight input-conditioned spectral gating that modulates the SCR refinement. 
Our analysis establishes the structural distinction between prediction-space and correction-space interaction. Controlled comparisons demonstrate the empirical benefits of this design choice, while parameter-matched controls and experiments across alternative adapters further support the effectiveness of \textsc{CoRe}.
Across seven backbones, six datasets, and four horizons, \textsc{CoRe} improves forecasting accuracy in most evaluated settings with modest computational overhead.

\FloatBarrier

\bibliographystyle{abbrvnat}
\bibliography{references}

\clearpage
\appendix
\makeatletter
\def\fps@table{H}
\makeatother
\FloatBarrier

\section{Extended Related Work}
\label{app:extended_related}

\paragraph{Cross-variate modeling.}
Multivariate forecasting approaches are commonly described as
channel-independent, channel-dependent, or partially coupled
architectures~\citep{qiu2025comprehensive}.
Channel-independent models avoid explicit cross-variate interaction,
whereas channel-dependent models couple variables through mechanisms such
as attention or frequency-domain mixing
~\citep{nie2023a,zeng2023transformers,zhou2021informer,
liu2024itransformer,yi2023frequency}.
Recent work also explores structured or selective interaction
~\citep{ekambaram2023tsmixer,han2024capacity,qiu2025duet}.
Empirical studies show that the effectiveness of cross-variate modeling
varies across data characteristics
~\citep{toner2024analysis,chen2024calibration,
he2025robust,lin2025cats}.

\paragraph{Distribution shift and adaptation.}
Normalization-based methods such as RevIN, Non-Stationary Transformer,
Dish-TS, SAN, and FAN modify inputs or representations to improve
robustness to non-stationarity
~\citep{kim2021reversible,liu2022non,fan2023dish,liu2023adaptive,ye2024frequency}.
Other work studies online learning~\citep{wen2023onenet,zhang2024addressing}, domain adaptation~\citep{jin2022domain},  domain generalization~\citep{deng2026adaptive}, 
and test-time learning~\citep{christou2024test,medeiros2025accurate} under different supervision and optimization
settings.
\textsc{CoRe} focuses specifically on the interaction space for
cross-variate adaptation under delayed-supervision TSF-TTA.

\section{Implementation Details}
\label{app:implementation}

\subsection{Spectral Descriptor Computation}
\label{app:spectral_desc}

For an input window
$\mathbf{X}_t\in\mathbb{R}^{L\times C}$,
we first remove the temporal mean of each variate and compute its
one-sided power spectrum using the real-valued Fast Fourier Transform:
\begin{equation}
    P_t^{(c)}(f)
    =
    \left|
    \mathrm{rFFT}
    \left(
        X_t^{(c)}-\bar X_t^{(c)}
    \right)[f]
    \right|^2,
    \qquad
    f=0,\ldots,K_f-1,
\end{equation}
where
$K_f=\lfloor L/2\rfloor+1$.
Mean subtraction removes the DC component so that the descriptors
characterize temporal variation rather than the absolute level of the
series.

We average the spectra across variates and normalize them into a
distribution over frequency bins:
\begin{equation}
    P_t(f)
    =
    \frac{1}{C}\sum_{c=1}^{C}P_t^{(c)}(f),
    \qquad
    \bar P_t(f)
    =
    \frac{P_t(f)+\varepsilon}
    {\sum_{f'}P_t(f')+K_f\varepsilon}.
    \label{eq:app_spectral_norm}
\end{equation}
The smoothing constant $\varepsilon>0$ avoids undefined logarithms, and
$\sum_f\bar P_t(f)=1$.

\paragraph{Spectral entropy.}
We compute normalized spectral entropy as
\begin{equation}
    \mathrm{SE}_t
    =
    -
    \frac{
        \sum_{f=0}^{K_f-1}
        \bar P_t(f)\log\bar P_t(f)
    }
    {\log K_f}
    \in[0,1].
    \label{eq:app_se}
\end{equation}
Spectral entropy summarizes how concentrated or dispersed spectral energy
is across frequencies
~\citep{powell1979spectral,inouye1991quantification}.

\paragraph{Band-energy ratios.}
To complement spectral concentration with coarse information about where
the energy is located, we partition the frequency range into low, middle,
and high thirds:
\begin{align}
    \mathrm{LBR}_t
    &=
    \sum_{f=0}^{\lfloor K_f/3\rfloor}
    \bar P_t(f),\\
    \mathrm{MBR}_t
    &=
    \sum_{f=\lfloor K_f/3\rfloor+1}^{\lfloor2K_f/3\rfloor}
    \bar P_t(f),\\
    \mathrm{HBR}_t
    &=
    \sum_{f=\lfloor2K_f/3\rfloor+1}^{K_f-1}
    \bar P_t(f).
\end{align}
By construction,
\begin{equation}
    \mathrm{LBR}_t+\mathrm{MBR}_t+\mathrm{HBR}_t=1.
\end{equation}
The complete descriptor is
\begin{equation}
    \mathbf{s}_t
    =
    [
        \mathrm{SE}_t,
        \mathrm{LBR}_t,
        \mathrm{MBR}_t,
        \mathrm{HBR}_t
    ]
    \in\mathbb{R}^{4}.
\end{equation}
The entropy term describes spectral concentration, while the three band
ratios provide complementary information about coarse frequency
allocation.

\subsection{Full Online Adaptation Procedure}
\label{app:training}

Algorithm~\ref{alg:core_full} summarizes the complete streaming
procedure.
The forecasting backbone remains frozen throughout deployment.
Bias terms in the SCR projections are omitted for notational simplicity.
The base-adapter parameters $\phi$ and the parameters introduced by
\textsc{CoRe},
\begin{equation}
    \Theta
    =
    \{
        \mathbf W_{\downarrow},
        \mathbf W_{\uparrow},
        \mathbf W_g,
        \mathbf b_g
    \},
\end{equation}
are updated only using forecast--target pairs whose targets have already
been revealed.

At adaptation time $t$, let $\mathcal{B}_t$ denote the adaptation batch whose corresponding targets are available. This definition is identical to the leakage-free objective in Section~\ref{sec:objective}.

\begin{algorithm}[!ht]
\caption{\textsc{CoRe} Full Procedure}
\label{alg:core_full}
\DontPrintSemicolon
\SetKwInOut{Input}{Input}
\SetKwInOut{Output}{Output}

\Input{
test stream $\{\mathbf X_t\}$;
frozen backbone $f_\theta$;
base adapter $\mathcal A_\phi$;
\textsc{CoRe} parameters $\Theta$;
buffer size $B$;
adaptation steps $S$
}
\Output{adapted forecasts $\{\hat{\mathbf Y}_t\}$}

Initialize $\phi$ and $\Theta$\;
Initialize optional auxiliary state used by the base adapter\;

\For{each forecast time $t$}{
    $\hat{\mathbf Y}^{\mathrm{base}}_t
    \leftarrow f_\theta(\mathbf X_t)$\;

    Compute spectral descriptor
    $\mathbf s_t$ from $\mathbf X_t$\;

    $\mathbf g_t
    \leftarrow
    \tanh
    \left(
        \mathbf W_g\mathbf s_t+\mathbf b_g
    \right)$\;

    $\boldsymbol{\Delta}_t
    \leftarrow
    \mathcal A_\phi
    \left(
        \hat{\mathbf Y}^{\mathrm{base}}_t,
        \mathbf I_t
    \right)$\;

    $\boldsymbol{\alpha}_t
    \leftarrow
    \frac{1}{C}
    \sum_{c=1}^{C}
    \boldsymbol{\Delta}_t^{(c)}$\;

    \For{$c=1$ \KwTo $C$}{
        $\mathbf z_t^{(c)}
        \leftarrow
        [
            \boldsymbol{\Delta}_t^{(c)}
            \|
            \boldsymbol{\alpha}_t
        ]$\;

        $\boldsymbol{\delta}_t^{(c)}
        \leftarrow
        \mathbf W_{\uparrow}
        \tanh
        \left(
            \mathbf W_{\downarrow}
            \mathbf z_t^{(c)}
        \right)$\;
    }

    $\widetilde{\boldsymbol{\Delta}}_t^{(c)}
    \leftarrow
    \boldsymbol{\Delta}_t^{(c)}
    +
    g_t^{(c)}
    \boldsymbol{\delta}_t^{(c)}
    \quad\forall c$\;

    $\hat{\mathbf Y}_t
    \leftarrow
    \hat{\mathbf Y}^{\mathrm{base}}_t
    +
    \widetilde{\boldsymbol{\Delta}}_t$\;

    Form the adaptation batch $\mathcal{B}_t$ from samples whose corresponding
    targets are available\;

    Update the optional base-adapter state $\mathbf I_t$
    using only revealed observations\;

    \If{$|\mathcal B_t|>0$}{
        \For{$s=1$ \KwTo $S$}{
            $\mathcal L_t
            \leftarrow
            \displaystyle
            \frac{1}{|\mathcal B_t|}
            \sum_{\tau\in\mathcal B_t}
            \left\|
                \hat{\mathbf Y}_\tau-\mathbf Y_\tau
            \right\|_F^2
            +
            \lambda_{\mathrm{reg}}
            \left(
                \|\phi\|_2^2+\|\Theta\|_2^2
            \right)$\;

            Clip gradients to norm $1.0$\;

            Update $\phi$ and $\Theta$ according to the
            base-adapter optimization protocol\;
        }
    }
}
\end{algorithm}

\paragraph{Initialization.}\label{app:scr_init}
The SCR projection matrices
$\mathbf W_{\downarrow}$ and $\mathbf W_{\uparrow}$
use Xavier-uniform initialization with gain $0.01$ and zero biases.
The spectral projection $\mathbf W_g$ is initialized to zero and
$\mathbf b_g=-\mathbf 1$, giving
\begin{equation}
    g_t^{(c)}
    =
    \tanh(-1)
    \approx -0.76
\end{equation}
at initialization.
Thus, the initial SCR contribution is negatively modulated rather than
initialized at a large unconstrained value.

\paragraph{Hyperparameters.}
Unless otherwise specified, we follow the optimization protocol and
hyperparameters of the corresponding base adapter.
The main \textsc{CoRe}-specific architectural choice is the SCR bottleneck
rank $r$. We set $r=C$ on the six standard datasets. For the high-dimensional datasets (Electricity, Traffic) we use $r=16\ll C$, with discussion in Appendix~\ref{app:rank_ablations}.
The spectral-gate bias is initialized as
$\mathbf b_g=-\mathbf 1$.

\subsection{Baseline Configuration Details}
\label{app:baseline_config}
For reproducibility, we summarize the baseline configurations used in our experiments. Unless noted otherwise, we follow the default settings in the official repositories.

\paragraph{TAFAS ~\citep{kim2025battling}}
Following the official configuration,\footnote{\url{https://github.com/kimanki/TAFAS/config.py}} we employ OPTIMIZING\_METHOD = adam with BASE\_LR = 0.005, WEIGHT\_DECAY = 0.0001, MOMENTUM = 0.9, NESTEROV = True, and DAMPENING = 0.0. Method settings are PAAS = True, PERIOD\_N = 1, BATCH\_SIZE = 64, STEPS = 1, ADJUST\_PRED = True, CALI\_MODULE = True, GATING\_INIT = 0.01, HIDDEN\_DIM = 128, and GCM\_VAR\_WISE = True. The adapted module is the calibration module only (TTA.MODULE\_NAMES\_TO\_ADAPT = `cali').

\paragraph{COSA ~\citep{im2026cosa}}
Following the official configuration,\footnote{\url{https://github.com/bigbases/COSA\_ICLR2026/config.py}}
we use OPTIMIZING\_METHOD = adam with BASE\_LR = 0.005, WEIGHT\_DECAY = 0.0001, MOMENTUM = 0.9, NESTEROV = True, and DAMPENING = 0.0. Method settings are BATCH\_SIZE = 25, STEPS = 20, BUFFER\_SIZE = 10, BUFFER\_CONTEXT\_SIZE = 5, ADAPT\_FREQUENCY = 50, FAST\_ADAPTATION = True, ADAPTIVE\_LR = True, MAX\_LR = 0.005, MIN\_LR = 0.0001, MOMENTUM\_FACTOR = 0.9, CONVERGENCE\_THRESHOLD = 1e-4, VAR\_WISE\_GATING = True, ADAPTER\_LAYERS = 1, HIDDEN\_DIM = 64, PER\_BATCH\_LR\_RESET = True, PAAS = False, and PERIOD\_N = 1. The adapted module is the calibration module only (TTA.MODULE\_NAMES\_TO\_ADAPT = `cali').

We configure the adaptation steps in \textsc{CoRe} following the established settings of the base adapter, which we also empirically found to be optimal.

\subsection{Variance Across Random Seeds}
\label{app:seed_var}

Table~\ref{tab:std} reports the standard deviation of \ours{} across
10 random seeds for the 168 main experimental settings.
The observed variance is small relative to the average performance
differences reported in the main results.

\begin{table*}[!ht]
\centering
\caption{Standard deviation ($\pm1\sigma$, over 10 random seeds) of \ours{} MSE.}
\label{tab:std}
{\scriptsize
\setlength{\tabcolsep}{3.5pt}
\renewcommand{\arraystretch}{0.85}
\begin{adjustbox}{max width=\textwidth}
\begin{tabular}{ll cccc cccc cccc cccc}
\toprule
& & \multicolumn{4}{c}{\textbf{iTransformer}}
& \multicolumn{4}{c}{\textbf{DLinear}}
& \multicolumn{4}{c}{\textbf{FreTS}}
& \multicolumn{4}{c}{\textbf{Informer}} \\
\cmidrule(lr){3-6}\cmidrule(lr){7-10}\cmidrule(lr){11-14}\cmidrule(lr){15-18}
& $H$ & 96 & 192 & 336 & 720
      & 96 & 192 & 336 & 720
      & 96 & 192 & 336 & 720
      & 96 & 192 & 336 & 720 \\
\midrule
\multirow{6}{*}{\rotatebox[origin=c]{90}{\textbf{Dataset}}}
& ETTh1        & .0034 & .0036 & .0032 & .0024 & .0032 & .0066 & .0024 & .0015 & .0036 & .0050 & .0042 & .0041 & .0032 & .0048 & .0029 & .0023 \\
& ETTh2        & .0054 & .0088 & .0042 & .0033 & .0036 & .0036 & .0027 & .0019 & .0053 & .0074 & .0033 & .0024 & .0122 & .0115 & .0081 & .0039 \\
& ETTm1        & .0034 & .0041 & .0036 & .0040 & .0042 & .0093 & .0070 & .0093 & .0028 & .0032 & .0038 & .0053 & .0128 & .0055 & .0046 & .0047 \\
& ETTm2        & .0046 & .0043 & .0029 & .0011 & .0016 & .0020 & .0026 & .0031 & .0033 & .0025 & .0025 & .0027 & .0061 & .0034 & .0044 & .0030 \\
& Exchange     & .0047 & .0018 & .0023 & .0036 & .0041 & .0012 & .0028 & .0041 & .0033 & .0021 & .0024 & .0048 & .0021 & .0029 & .0096 & .0058 \\
& Weather      & .0020 & .0027 & .0024 & .0020 & .0040 & .0039 & .0024 & .0014 & .0025 & .0012 & .0025 & .0025 & .0012 & .0047 & .0028 & .0022 \\
\bottomrule
\end{tabular}
\end{adjustbox}

\vspace{4pt}

\begin{adjustbox}{max width=\textwidth}
\begin{tabular}{ll cccc cccc cccc}
\toprule
& & \multicolumn{4}{c}{\textbf{PatchTST}}
& \multicolumn{4}{c}{\textbf{OLS}}
& \multicolumn{4}{c}{\textbf{MICN}} \\
\cmidrule(lr){3-6}\cmidrule(lr){7-10}\cmidrule(lr){11-14}
& $H$ & 96 & 192 & 336 & 720
      & 96 & 192 & 336 & 720
      & 96 & 192 & 336 & 720 \\
\midrule
\multirow{6}{*}{\rotatebox[origin=c]{90}{\textbf{Dataset}}}
& ETTh1        & .0021 & .0040 & .0019 & .0038 & .0021 & .0040 & .0031 & .0026 & .0055 & .0037 & .0036 & .0028 \\
& ETTh2        & .0033 & .0053 & .0031 & .0017 & .0044 & .0048 & .0028 & .0024 & .0054 & .0024 & .0027 & .0031 \\
& ETTm1        & .0033 & .0046 & .0041 & .0052 & .0047 & .0130 & .0050 & .0054 & .0041 & .0043 & .0055 & .0116 \\
& ETTm2        & .0023 & .0014 & .0017 & .0015 & .0029 & .0019 & .0032 & .0036 & .0033 & .0020 & .0023 & .0026 \\
& Exchange     & .0032 & .0025 & .0023 & .0019 & .0023 & .0020 & .0024 & .0039 & .0037 & .0008 & .0030 & .0046 \\
& Weather      & .0021 & .0030 & .0019 & .0016 & .0043 & .0030 & .0027 & .0022 & .0031 & .0020 & .0031 & .0012 \\
\bottomrule
\end{tabular}
\end{adjustbox}
}
\end{table*}

\FloatBarrier

\section{Additional Analysis of Correction-Space Interaction}
\label{app:additionalexp}

\subsection{Full Correction-Space vs.\ Prediction-Space Comparison}
\label{app:corrvspred}

\begin{figure}[t]
    \centering
    \includegraphics[width=0.85\linewidth]
    {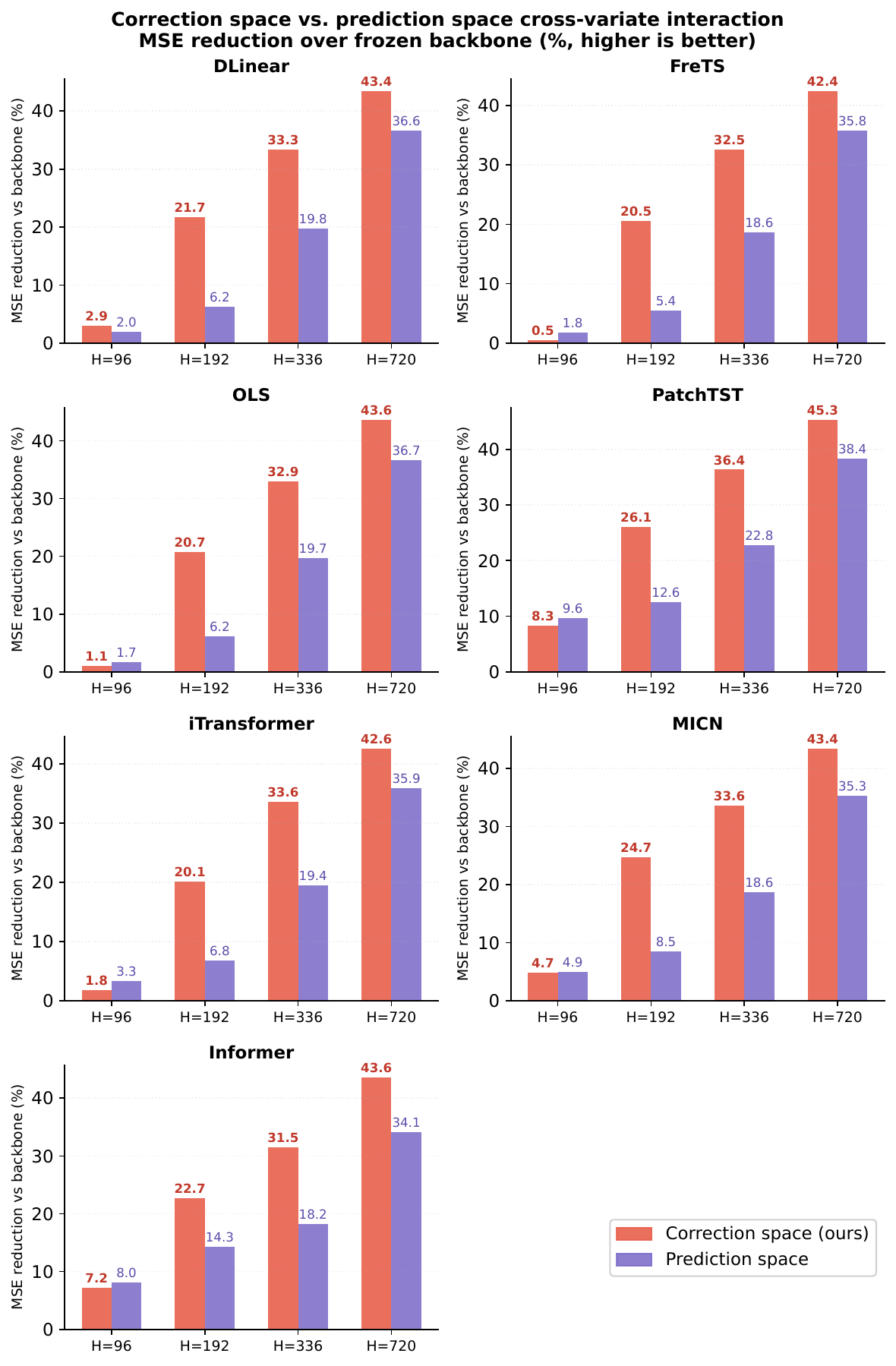}
    \caption{
    Controlled comparison of correction-space and prediction-space
    interaction across all seven backbones.
    }
    \label{fig:corrpred_all_backbones}
\end{figure}

We provide the full comparison between correction-space and prediction-space interaction across all seven backbones.
This controlled experiment isolates the interaction space while keeping
the refinement architecture and base-adapter correction unchanged.
Figure~\ref{fig:corrpred_all_backbones} reports the resulting MSE
reductions.

At $H=96$, the two interaction schemes are generally comparable.
The separation becomes clearer at medium-to-long horizons, where
correction-space interaction obtains larger MSE reductions across the
evaluated backbones.
At $H=192$, $H=336$, and $H=720$, the largest observed margins between
the two variants are 16.2\%, 15.0\%, and 9.5\%, respectively.
Together with Proposition~\ref{prop:structure}, these results show that
the choice of interaction space has a substantial empirical consequence
across backbone architectures.

\subsection{Generality across Base Adapters}
\label{app:alternative_adapters}

We examine whether correction-space interaction remains effective beyond
the COSA base adapter used in the main experiments.
We evaluate \textsc{CoRe} with two structurally different correction
interfaces: TAFAS and a standalone MLP adapter.
Together with COSA, these experiments cover a context-conditioned
adapter, a gated calibration module, and a minimal additive correction
adapter.

\subsubsection{\textsc{CoRe} on TAFAS}
\label{app:tafas}

TAFAS applies an output Gated Calibration Module (GCM):
\begin{equation}
\label{eq:gcm_correction_space}
\mathrm{GCM}(\hat{\mathbf Y}_t)
=
\hat{\mathbf Y}_t
+
\tanh(\boldsymbol{\alpha})
\circ
\underbrace{
\left(
\{\mathbf W^c\hat{\mathbf Y}_t^c\}_{c=1}^{C}
+
\mathbf b
\right)
}_{\text{correction}},
\end{equation}
where
$\mathbf W^c\in\mathbb R^{H\times H}$,
$\mathbf b\in\mathbb R^{H\times C}$,
and
$\boldsymbol{\alpha}\in\mathbb R^C$.

We apply SCR to the additive correction inside the GCM before its
original multiplicative gate is applied.
Across six datasets and four horizons using DLinear,
this yields a 4.49\% average MSE reduction over TAFAS
(Table~\ref{tab:tafas_core}).

\begin{table}[t]
\centering
\caption{MSE of TAFAS and TAFAS+\ours{} (ours). \textbf{Bold} = better result. Avg.\ Gain = relative MSE reduction averaged over $H \in \{96,192,336,720\}$.}
\label{tab:tafas_core}
\resizebox{\textwidth}{!}{
\begin{tabular}{llccccccccr}
\toprule
\multirow{2}{*}{Backbone} & \multirow{2}{*}{Dataset}
  & \multicolumn{2}{c}{$H=96$} & \multicolumn{2}{c}{$H=192$}
  & \multicolumn{2}{c}{$H=336$} & \multicolumn{2}{c}{$H=720$}
  & \multirow{2}{*}{Avg.\ Gain} \\
\cmidrule(lr){3-4}\cmidrule(lr){5-6}\cmidrule(lr){7-8}\cmidrule(lr){9-10}
& & TAFAS & +CoRe & TAFAS & +\ours{} & TAFAS & +\ours{} & TAFAS & +\ours{} & \\
\midrule
\multirow{6}{*}{DLinear}
  & ETTh1    & 0.4604 & \textbf{0.4604 }    & 0.5099 & \textbf{0.5020} & 0.5622 & \textbf{0.5515} & 0.6685 & \textbf{0.6548} & $+$1.83\% \\
  & ETTh2    & 0.2302 & \textbf{0.2039} & 0.2836 & \textbf{0.2710} & \textbf{0.3193} & 0.3229          & \textbf{0.3939} & 0.4140          & $+$2.41\% \\
  & ETTm1    & 0.3488 & \textbf{0.3342} & 0.4159 & \textbf{0.3924} & 0.4787 & \textbf{0.4681} & \textbf{0.5495} & 0.5539          & $+$2.81\% \\
  & ETTm2    & 0.1588 & \textbf{0.1505} & 0.1927 & \textbf{0.1841} & 0.2337 & \textbf{0.2298} & 0.3053 & \textbf{0.3008} & $+$3.21\% \\
  & Weather  & 0.1828 & \textbf{0.1531} & 0.2219 & \textbf{0.2136} & 0.2695 & \textbf{0.2454} & 0.3550 & \textbf{0.3164} & $+$9.95\% \\
  & Exchange & 0.0878 & \textbf{0.0789} & 0.1736 & \textbf{0.1599} & 0.3067 & \textbf{0.2820} & 0.8281 & \textbf{0.8217} & $+$6.71\% \\
\cmidrule(lr){2-11}
  & \textit{Avg.} & \multicolumn{8}{c}{} & $+$4.49\% \\
\bottomrule
\end{tabular}}
\end{table}

\subsubsection{\textsc{CoRe} on a Standalone MLP Adapter}
\label{app:core_mlp}

We also construct a minimal standalone correction adapter:
\begin{equation}
    \boldsymbol{\Delta}_t
    =
    \mathrm{MLP}_{\mathrm{base}}
    \left(
        \hat{\mathbf Y}^{\mathrm{base}}_t
    \right)
    \in\mathbb R^{H\times C}.
\end{equation}
The base adapter is a two-layer MLP with GELU activation and a
zero-initialized output layer.
It contains no context buffer, adaptive learning-rate mechanism, or
periodicity-aware scheduling.

SCR then operates on
$\boldsymbol{\Delta}_t$ exactly as in Section~\ref{sec:scr}.
Across seven backbones, six datasets, and four horizons,
the resulting method improves 156/168 settings over the standalone
adapter, with a 3.70\% average MSE reduction
(Table~\ref{tab:coresa}).
The remaining negative differences are small in absolute magnitude
($|\Delta\mathrm{MSE}|<0.005$).

\begin{table}[H]
\centering
\small
\setlength{\tabcolsep}{2.5pt}
\caption{MSE of \ours{}-MLP without and with SCR, averaged over
4 horizons and 10 seeds. \textbf{w/o SCR}: standalone MLP
correction only. \textbf{\ours{}-MLP}: MLP correction + SCR.
Gain: relative MSE reduction from SCR.}
\label{tab:coresa}
\resizebox{\linewidth}{!}{%
\begin{tabular}{lcccccccc}
\toprule
 & \multicolumn{2}{c}{ETTh1} & \multicolumn{2}{c}{ETTh2} &
   \multicolumn{2}{c}{ETTm1} & \multicolumn{2}{c}{ETTm2} \\
\cmidrule(lr){2-3}\cmidrule(lr){4-5}\cmidrule(lr){6-7}\cmidrule(lr){8-9}
Backbone & w/o SCR & \ours{}-MLP & w/o SCR & \ours{}-MLP & w/o SCR & \ours{}-MLP & w/o SCR & \ours{}-MLP\\
\midrule
DLinear     & 0.4842 & \textbf{0.4767} & 0.2519 & \textbf{0.2378} & 0.5285 & \textbf{0.5101} & 0.2096 & \textbf{0.1949} \\
FreTS       & 0.4722 & \textbf{0.4681} & 0.2556 & \textbf{0.2439} & 0.5243 & \textbf{0.4957} & 0.2056 & \textbf{0.1923} \\
iTransformer& 0.4685 & \textbf{0.4673} & 0.2710 & \textbf{0.2605} & 0.5337 & \textbf{0.5145} & 0.2296 & \textbf{0.2131} \\
PatchTST    & 0.4595 & \textbf{0.4526} & 0.2534 & \textbf{0.2494} & 0.5265 & \textbf{0.5038} & 0.2121 & \textbf{0.1980} \\
OLS         & 0.4706 & \textbf{0.4661} & 0.2503 & \textbf{0.2383} & 0.5270 & \textbf{0.5087} & 0.2086 & \textbf{0.1966} \\
MICN        & 0.5486 & \textbf{0.5420} & 0.2697 & \textbf{0.2601} & 0.5561 & \textbf{0.5443} & 0.2124 & \textbf{0.2017} \\
Informer    & 0.6207 & \textbf{0.6180} & 0.3841 & \textbf{0.3768} & 0.6913 & \textbf{0.6642} & 0.2618 & \textbf{0.2533} \\
\midrule
 & \multicolumn{2}{c}{Exchange} & \multicolumn{2}{c}{Weather} &
   \multicolumn{2}{c}{Avg. MSE} & \multicolumn{2}{c}{Gain / $n$} \\
\cmidrule(lr){2-3}\cmidrule(lr){4-5}\cmidrule(lr){6-7}\cmidrule(lr){8-9}
Backbone & w/o SCR & \ours{}-MLP & w/o SCR & \ours{}-MLP & w/o SCR & \ours{}-MLP & & \\
\midrule
DLinear     & 0.1087 & \textbf{0.0978} & 0.2124 & \textbf{0.2009} & 0.2992 & \textbf{0.2864} & +4.28\% & 23/24 \\
FreTS       & 0.1161 & \textbf{0.1050} & 0.2077 & \textbf{0.1972} & 0.2969 & \textbf{0.2838} & +4.41\% & 22/24 \\
iTransformer& 0.1549 & \textbf{0.1354} & 0.2037 & \textbf{0.1929} & 0.3102 & \textbf{0.2973} & +4.16\% & 23/24 \\
PatchTST    & 0.1178 & \textbf{0.1122} & 0.2022 & \textbf{0.1896} & 0.2953 & \textbf{0.2843} & +3.73\% & 22/24 \\
OLS         & 0.1071 & \textbf{0.1026} & 0.2136 & \textbf{0.1993} & 0.2962 & \textbf{0.2853} & +3.68\% & 23/24 \\
MICN        & 0.1498 & \textbf{0.1362} & 0.1955 & \textbf{0.1870} & 0.3053 & \textbf{0.2952} & +3.31\% & 22/24 \\
Informer    & \textbf{0.2405} & 0.2427 & 0.2629 & \textbf{0.2491} & 0.4102 & \textbf{0.4007} & +2.32\% & 21/24 \\
\bottomrule
\end{tabular}%
}
\end{table}

Adding SCR to TAFAS and to a standalone two-layer MLP adapter reduces MSE
by 4.49\% and 3.70\% on average, respectively.
The gains across two distinct base-adapter designs support
correction-space interaction beyond the particular COSA instantiation
used in the main experiments.

\subsection{Backbone-Error Diagnostics}
\label{app:backbone_quality}

We examine whether the correction signal underlying correction-space
interaction varies systematically with frozen-backbone error.
Across all 168 main settings, we relate the frozen-backbone error $e_t$
for each test window to (i) the correction magnitude
$\|\boldsymbol{\Delta}_t\|$ and (ii) the relative improvement
\begin{equation}
    \mathrm{RI}_t
    =
    \frac{
        e_t-e_t^{\textsc{CoRe}}
    }{e_t}.
\end{equation}

Across settings, backbone error is positively associated with correction
magnitude, with mean Pearson correlation $r=0.59$ and positive
correlation in 167/168 settings.
After normalization by local signal scale, the error--correction
association remains similar ($r=0.60$).

Backbone error is also positively associated with relative improvement,
with mean $r=0.34$ and positive correlation in 166/168 settings.
Restricting the analysis to non-overlapping windows gives mean
$r=0.36$, positive in 138/161 measurable settings; seven Exchange-Rate
settings at $H=720$ contain too few non-overlapping windows for reliable
estimation.

Overall, larger frozen-backbone errors are associated with stronger
learned corrections across nearly all settings, showing that the
correction signal varies systematically with the amount of
frozen-backbone error observed in the stream.

\section{Additional Comparisons and Generalization Results}
\label{app:comparisons_generalization}

\subsection{Comparison with Other TSF-TTA Methods}
\label{app:tsf_tta_methods}
\paragraph{PETSA.}\label{app:petsa}
PETSA~\citep{medeiros2025accurate} is a parameter-efficient TSF-TTA method that uses low-rank, dynamically gated calibration modules at its input and output.

We compare \textsc{CoRe} with PETSA using the same frozen backbone checkpoints and delayed-supervision forecasting protocol, providing a controlled comparison with a recent TSF-TTA method.

Table~\ref{tab:petsa_results} reports the full comparison across all 168
settings (seven backbones $\times$ six datasets $\times$ four horizons).
\textsc{CoRe} reduces MSE relative to PETSA by 23.39\% on average and
outperforms it in 145/168 settings.
The average gain is positive for all seven backbones, ranging from
20.86\% on iTransformer to 27.74\% on Informer.

\begin{table}[H]
\centering
\caption{Comparison with PETSA~\citep{medeiros2025accurate}. We report the MSE reduction of \ours{} relative to PETSA for each backbone, averaged over six datasets and four prediction horizons. Both methods use the same frozen backbone checkpoints. Higher is better.}
\label{tab:petsa_results}
\small
\begin{tabular}{lc}
\toprule
\textbf{Backbone} & \textbf{\ours{} vs.\ PETSA} \\
\midrule
DLinear      & $22.64\%$ \\
FreTS        & $21.90\%$ \\
iTransformer & $20.86\%$ \\
MICN         & $24.83\%$ \\
OLS          & $21.71\%$ \\
PatchTST     & $24.03\%$ \\
Informer     & $27.74\%$ \\
\midrule
\textbf{Average} & $\mathbf{23.39\%}$ \\
\bottomrule
\end{tabular}
\end{table}

\paragraph{DynaTTA.}\label{app:dynatta}

We additionally compare with DynaTTA~\citep{grover2025shift}.
Following the evaluation protocol of \citet{im2026cosa}, we re-implement
DynaTTA within the common forecasting codebase for a consistent comparison.

The selected configuration uses
ALPHA\_MIN = $10^{-4}$,
ALPHA\_MAX = $10^{-3}$,
KAPPA = 1.0,
ETA = 0.1,
EPS = $10^{-6}$,
WARMUP\_FACTOR = 1,
MSE\_BUFFER\_SIZE = 256,
RTAB\_SIZE = 360,
RDB\_SIZE = 100,
METRIC\_HISTORY\_SIZE = 256,
UPDATE\_BUFFERS\_INTERVAL = 1,
and UPDATE\_METRICS\_INTERVAL = 1.

Table~\ref{tab:dynatta_results} reports the comparison across the
evaluated settings.
\textsc{CoRe} outperforms DynaTTA and COSA in the majority of cases,
with particularly large differences in several Exchange-Rate and Weather
settings.

\begin{table*}[!t]
\centering
\caption{Prediction MSE comparison with DynaTTA across five backbones (lower is better). \textbf{Bold}: best among TTA methods. \underline{Underline}: second-best among TTA methods. DynaTTA results use the COSA re-implementation with prediction-drift signal.}
\label{tab:dynatta_results}
\small
\setlength{\tabcolsep}{1pt}
\renewcommand{\arraystretch}{0.9}
\resizebox{\textwidth}{!}{%
\begin{tabular}{ll ccc ccc ccc ccc ccc}
\toprule
& & \multicolumn{3}{c}{\textbf{DLinear}} & \multicolumn{3}{c}{\textbf{FreTS}} & \multicolumn{3}{c}{\textbf{OLS}} & \multicolumn{3}{c}{\textbf{iTransformer}} & \multicolumn{3}{c}{\textbf{MICN}} \\
\cmidrule(lr){3-5}\cmidrule(lr){6-8}\cmidrule(lr){9-11}\cmidrule(lr){12-14}\cmidrule(lr){15-17}
& $H$ & DynaTTA & COSA & \ours{} & DynaTTA & COSA & \ours{} & DynaTTA & COSA & \ours{} & DynaTTA & COSA & \ours{} & DynaTTA & COSA & \ours{} \\
\midrule
\multirow{4}{*}{\rotatebox[origin=c]{90}{\textbf{ETTh1}}}
& 96  & \textbf{0.4708} & 0.4922 & \underline{0.4916} & \textbf{0.4511} & \underline{0.4623} & 0.4625 & \textbf{0.4486} & 0.4729 & \underline{0.4723} & \textbf{0.4509} & 0.4638 & \underline{0.4545} & \textbf{0.5804} & \underline{0.5837} & 0.5863 \\
& 192 & \underline{0.5321} & 0.5371 & \textbf{0.4961} & 0.5138 & \underline{0.5084} & \textbf{0.4692} & \underline{0.5082} & 0.5173 & \textbf{0.4763} & 0.5156 & \underline{0.5024} & \textbf{0.4589} & 0.6032 & \underline{0.5802} & \textbf{0.5203} \\
& 336 & 0.5792 & \underline{0.5208} & \textbf{0.4722} & 0.5838 & \underline{0.4970} & \textbf{0.4596} & 0.5626 & \underline{0.5041} & \textbf{0.4568} & 0.6052 & \underline{0.4956} & \textbf{0.4661} & 0.7037 & \underline{0.5889} & \textbf{0.5386} \\
& 720 & 0.7047 & \underline{0.5433} & \textbf{0.4891} & 0.7086 & \underline{0.5616} & \textbf{0.5052} & 0.6933 & \underline{0.5443} & \textbf{0.4886} & 0.7118 & \underline{0.5381} & \textbf{0.5055} & 0.8282 & \underline{0.6177} & \textbf{0.5591} \\
\midrule
\multirow{4}{*}{\rotatebox[origin=c]{90}{\textbf{ETTh2}}}
& 96  & \textbf{0.2338} & 0.2552 & \underline{0.2426} & \textbf{0.2397} & 0.2560 & \underline{0.2462} & \textbf{0.2325} & 0.2443 & \underline{0.2389} & \textbf{0.2767} & 0.3050 & \underline{0.2803} & 0.2612 & \underline{0.2538} & \textbf{0.2476} \\
& 192 & 0.2888 & \underline{0.2739} & \textbf{0.2275} & 0.2915 & \underline{0.2676} & \textbf{0.2371} & 0.2909 & \underline{0.2740} & \textbf{0.2290} & \underline{0.3077} & 0.3116 & \textbf{0.2722} & 0.3454 & \underline{0.2923} & \textbf{0.2430} \\
& 336 & 0.3380 & \underline{0.2521} & \textbf{0.2171} & 0.3476 & \underline{0.2548} & \textbf{0.2196} & 0.3428 & \underline{0.2448} & \textbf{0.2122} & 0.3703 & \underline{0.2552} & \textbf{0.2245} & 0.4000 & \underline{0.2828} & \textbf{0.2486} \\
& 720 & 0.4360 & \underline{0.2589} & \textbf{0.2324} & 0.4310 & \underline{0.2516} & \textbf{0.2336} & 0.4273 & \underline{0.2499} & \textbf{0.2301} & 0.4574 & \underline{0.2695} & \textbf{0.2545} & 0.5259 & \underline{0.3058} & \textbf{0.2746} \\
\midrule
\multirow{4}{*}{\rotatebox[origin=c]{90}{\textbf{ETTm1}}}
& 96  & \underline{0.3814} & \textbf{0.3800} & 0.3881 & \textbf{0.3731} & \underline{0.3836} & 0.3862 & \underline{0.3865} & \textbf{0.3780} & 0.3874 & \textbf{0.3694} & 0.3799 & \underline{0.3776} & \textbf{0.4186} & 0.4240 & \underline{0.4232} \\
& 192 & 0.4809 & \underline{0.4755} & \textbf{0.4282} & 0.4826 & \underline{0.4615} & \textbf{0.4200} & 0.4786 & \underline{0.4730} & \textbf{0.4235} & 0.4790 & \underline{0.4346} & \textbf{0.4184} & 0.5433 & \underline{0.4822} & \textbf{0.4551} \\
& 336 & 0.5711 & \underline{0.5358} & \textbf{0.4665} & 0.5818 & \underline{0.5195} & \textbf{0.4616} & 0.5677 & \underline{0.5360} & \textbf{0.4658} & 0.5760 & \underline{0.4825} & \textbf{0.4585} & 0.6056 & \underline{0.5480} & \textbf{0.5018} \\
& 720 & 0.6271 & \underline{0.5920} & \textbf{0.5424} & 0.6903 & \underline{0.5773} & \textbf{0.5200} & 0.6557 & \underline{0.5938} & \textbf{0.5367} & 0.7122 & \underline{0.5615} & \textbf{0.5171} & 0.6977 & \underline{0.6061} & \textbf{0.5602} \\
\midrule
\multirow{4}{*}{\rotatebox[origin=c]{90}{\textbf{ETTm2}}}
& 96  & \textbf{0.1640} & 0.1915 & \underline{0.1644} & \textbf{0.1667} & 0.1900 & \underline{0.1707} & \underline{0.1683} & 0.1923 & \textbf{0.1655} & \underline{0.2149} & 0.2226 & \textbf{0.1831} & 0.2027 & \underline{0.1909} & \textbf{0.1845} \\
& 192 & 0.2040 & \underline{0.1961} & \textbf{0.1652} & 0.2069 & \underline{0.1885} & \textbf{0.1628} & 0.2113 & \underline{0.1978} & \textbf{0.1671} & 0.3044 & \underline{0.2393} & \textbf{0.1905} & 0.2471 & \underline{0.2307} & \textbf{0.1832} \\
& 336 & 0.2908 & \underline{0.1969} & \textbf{0.1725} & 0.2743 & \underline{0.1944} & \textbf{0.1686} & 0.2914 & \underline{0.1985} & \textbf{0.1726} & 0.4027 & \underline{0.2473} & \textbf{0.2058} & 0.3212 & \underline{0.2238} & \textbf{0.1871} \\
& 720 & 0.3771 & \underline{0.2066} & \textbf{0.1778} & 0.4174 & \underline{0.2110} & \textbf{0.1799} & 0.4574 & \underline{0.2067} & \textbf{0.1787} & 0.4922 & \underline{0.2329} & \textbf{0.1955} & 0.4980 & \underline{0.2381} & \textbf{0.1991} \\
\midrule
\multirow{4}{*}{\rotatebox[origin=c]{90}{\textbf{Exchange}}}
& 96  & 0.0948 & \underline{0.0816} & \textbf{0.0706} & 0.0907 & \underline{0.0835} & \textbf{0.0738} & 0.0918 & \underline{0.0795} & \textbf{0.0741} & 0.1015 & \underline{0.0913} & \textbf{0.0832} & 0.1207 & \underline{0.0920} & \textbf{0.0779} \\
& 192 & 0.1975 & \underline{0.0844} & \textbf{0.0733} & 0.1868 & \underline{0.0959} & \textbf{0.0806} & 0.1748 & \underline{0.0953} & \textbf{0.0812} & 0.2375 & \underline{0.1093} & \textbf{0.0988} & 0.2070 & \underline{0.0855} & \textbf{0.0739} \\
& 336 & 0.3001 & \underline{0.0867} & \textbf{0.0804} & 0.3108 & \underline{0.0978} & \textbf{0.0933} & 0.3024 & \underline{0.1003} & \textbf{0.0947} & 0.3328 & \underline{0.1192} & \textbf{0.1054} & 0.3579 & \underline{0.1299} & \textbf{0.1091} \\
& 720 & 0.8364 & \underline{0.1469} & \textbf{0.1239} & 0.8253 & \underline{0.1537} & \textbf{0.1289} & 0.8287 & \underline{0.1490} & \textbf{0.1259} & 0.9572 & \underline{0.2228} & \textbf{0.1957} & 0.8618 & \underline{0.2320} & \textbf{0.2211} \\
\midrule
\multirow{4}{*}{\rotatebox[origin=c]{90}{\textbf{Weather}}}
& 96  & 0.1950 & \underline{0.1814} & \textbf{0.1719} & 0.1865 & \underline{0.1739} & \textbf{0.1617} & 0.1993 & \underline{0.1812} & \textbf{0.1704} & \underline{0.1738} & 0.1753 & \textbf{0.1635} & 0.2105 & \underline{0.1701} & \textbf{0.1646} \\
& 192 & 0.3311 & \underline{0.1970} & \textbf{0.1711} & 0.2785 & \underline{0.1944} & \textbf{0.1690} & 0.3054 & \underline{0.1960} & \textbf{0.1719} & 0.2683 & \underline{0.2000} & \textbf{0.1697} & 0.3536 & \underline{0.2078} & \textbf{0.1608} \\
& 336 & 0.4103 & 0.2465 & \textbf{0.1782} & 0.3846 & 0.2341 & \textbf{0.1785} & 0.5927 & 0.2440 & \textbf{0.1777} & 0.4036 & \underline{0.2439} & \textbf{0.1736} & 0.5471 & \underline{0.2107} & \textbf{0.1698} \\
& 720 & 0.5105 & 0.2231 & \textbf{0.1777} & 0.5684 & 0.2084 & \textbf{0.1763} & 0.4995 & 0.2252 & \textbf{0.1785} & 0.4934 & \underline{0.2193} & \textbf{0.1876} & 0.6519 & \underline{0.2013} & \textbf{0.1711} \\
\bottomrule
\end{tabular}}
\end{table*}

\subsection{Comparison with Normalization-Based Methods}
\label{app:normal}

Table~\ref{tab:dlinear_revin_fan_core} compares \textsc{CoRe} with
RevIN~\citep{kim2021reversible}, which applies reversible instance
normalization, and FAN~\citep{ye2024frequency}, which uses
frequency-aware normalization, using DLinear.
RevIN and FAN improve several forecasting settings, while \textsc{CoRe}
achieves stronger aggregate performance under the evaluated test-time
adaptation protocol.

\begin{table*}[!ht]
\centering
\caption{Prediction MSE comparison with RevIN and FAN on DLinear (lower is better). }
\label{tab:dlinear_revin_fan_core}
\small
\setlength{\tabcolsep}{4pt}
\renewcommand{\arraystretch}{0.92}
\begin{tabular}{ccccc}
\toprule
Dataset & $H$ & RevIN & FAN & \textsc{CoRe} \\
\midrule

\multirow{4}{*}{{ETTh1}} & 96  & \textbf{0.4591} & 0.4620 & 0.4916 \\
  & 192 & 0.5121 & 0.5159 & \textbf{0.4961} \\
  & 336 & 0.5587 & 0.5427 & \textbf{0.4722} \\
  & 720 & 0.7063 & 0.6593 & \textbf{0.4891} \\
\midrule
\multirow{4}{*}{{ETTh2}} & 96  & \textbf{0.2302} & 0.2512 & 0.2426 \\
  & 192 & 0.2834 & 0.2973 & \textbf{0.2275} \\
  & 336 & 0.3228 & 0.3249 & \textbf{0.2171} \\
  & 720 & 0.4111 & 0.4188 & \textbf{0.2324} \\
\midrule

\multirow{4}{*}{{{ETTm1}}} & 96  & 0.3704 & \textbf{0.3695} & 0.3881 \\
  & 192 & 0.4431 & 0.8090 & \textbf{0.4282} \\
  & 336 & 0.5167 & 0.6710 & \textbf{0.4665} \\
  & 720 & 0.5904 & 0.5628 & \textbf{0.5424} \\
\midrule

\multirow{4}{*}{{{ETTm2}}} & 96  & \textbf{0.1599} & 0.1835 & 0.1644 \\
  & 192 & 0.1934 & 0.2436 & \textbf{0.1652} \\
  & 336 & 0.2326 & 0.2869 & \textbf{0.1725} \\
  & 720 & 0.3061 & 0.3395 & \textbf{0.1778} \\
\midrule

\multirow{4}{*}{{{Exchange}}} & 96  & 0.0864 & 0.0968 & \textbf{0.0706} \\
  & 192 & 0.1779 & 0.2079 & \textbf{0.0733} \\
  & 336 & 0.3257 & 0.3718 & \textbf{0.0804} \\
  & 720 & 0.8461 & 0.8001 & \textbf{0.1239} \\
\midrule
\multirow{4}{*}{{{Weather}}} & 96  & 0.1933 & 0.1827 & \textbf{0.1719} \\
  & 192 & 0.2371 & 0.2235 & \textbf{0.1711} \\
  & 336 & 0.2881 & 0.2705 & \textbf{0.1782} \\
  & 720 & 0.3598 & 0.3402 & \textbf{0.1777} \\

\bottomrule
\end{tabular}
\end{table*}

\subsection{Generalization to an Additional Backbone: TimeXer}
\label{app:timexer}

We further evaluate \textsc{CoRe} on another forecasting backbone, TimeXer~\citep{wang2024timexer} to
test whether correction-space interaction remains effective with a modern
backbone that already models cross-variate dependencies.
We integrate the official TimeXer implementation and evaluate the same
six datasets and four prediction horizons as in the main experiments.
Architecture hyperparameters follow the official configuration for each
dataset and horizon.
Because no official Exchange-Rate configuration is provided, we use the
ETTm2 configuration, which is closest in dimensionality.

Table~\ref{tab:timexer_results} compares frozen TimeXer with TAFAS,
PETSA, COSA, and \textsc{CoRe}.
All TTA results are averaged over 10 seeds.
Across six datasets and four horizons, \textsc{CoRe} improves over COSA in
23 of 24 settings, reducing MSE by 9.13\% on average.
It also reduces average MSE by 24.42\% relative to frozen TimeXer and
23.62\% relative to PETSA.
The only regression relative to COSA occurs on ETTm1 at $H=96$.

\begin{table*}[t]
\centering
\caption{Evaluation on TimeXer~\citep{wang2024timexer}. MSE is reported for the frozen TimeXer backbone and four TTA methods over six datasets and four horizons. TTA results are mean $\pm$ standard deviation over 10 seeds. Lower is better. Bold denotes the best TTA result in each setting.}
\label{tab:timexer_results}
\scriptsize
\setlength{\tabcolsep}{4pt}
\begin{tabular}{llccccc}
\toprule
\textbf{Dataset} & \textbf{$H$} & \textbf{Base (TimeXer)} & \textbf{TAFAS} & \textbf{COSA} & \textbf{\ours{}} & \textbf{PETSA} \\
\midrule
\multirow{4}{*}{ETTh1}
& 96  & 0.4301 & \textbf{0.4304$\pm$0.0004} & 0.4407$\pm$0.0001 & 0.4367$\pm$0.0019 & 0.4349$\pm$0.0002 \\
& 192 & 0.4895 & 0.4996$\pm$0.0001 & 0.4867$\pm$0.0004 & \textbf{0.4513$\pm$0.0049} & 0.4922$\pm$0.0003 \\
& 336 & 0.5605 & 0.5795$\pm$0.0113 & 0.4776$\pm$0.0003 & \textbf{0.4329$\pm$0.0025} & 0.5544$\pm$0.0005 \\
& 720 & 0.6931 & 0.6810$\pm$0.0010 & 0.5195$\pm$0.0001 & \textbf{0.4813$\pm$0.0046} & 0.6698$\pm$0.0014 \\
\midrule
\multirow{4}{*}{ETTh2}
& 96  & 0.2367 & \textbf{0.2317$\pm$0.0001} & 0.2492$\pm$0.0008 & 0.2469$\pm$0.0034 & 0.2367$\pm$0.0003 \\
& 192 & 0.3000 & 0.3021$\pm$0.0001 & 0.2907$\pm$0.0035 & \textbf{0.2420$\pm$0.0039} & 0.2986$\pm$0.0002 \\
& 336 & 0.3223 & 0.3254$\pm$0.0007 & 0.2494$\pm$0.0022 & \textbf{0.2261$\pm$0.0026} & 0.3166$\pm$0.0003 \\
& 720 & 0.4432 & 0.4460$\pm$0.0030 & 0.2701$\pm$0.0007 & \textbf{0.2446$\pm$0.0035} & 0.4135$\pm$0.0003 \\
\midrule
\multirow{4}{*}{ETTm1}
& 96  & 0.3979 & \textbf{0.3546$\pm$0.0008} & 0.3728$\pm$0.0010 & 0.3783$\pm$0.0039 & 0.3991$\pm$0.0015 \\
& 192 & 0.4345 & 0.4080$\pm$0.0001 & 0.4314$\pm$0.0002 & \textbf{0.4061$\pm$0.0045} & 0.4307$\pm$0.0004 \\
& 336 & 0.4890 & 0.4704$\pm$0.0003 & 0.4701$\pm$0.0030 & \textbf{0.4378$\pm$0.0043} & 0.4715$\pm$0.0005 \\
& 720 & 0.5525 & 0.5547$\pm$0.0013 & 0.5433$\pm$0.0009 & \textbf{0.4966$\pm$0.0143} & 0.5321$\pm$0.0007 \\
\midrule
\multirow{4}{*}{ETTm2}
& 96  & 0.1479 & \textbf{0.1476$\pm$0.0003} & 0.1750$\pm$0.0007 & 0.1580$\pm$0.0021 & 0.1506$\pm$0.0001 \\
& 192 & 0.1818 & 0.1853$\pm$0.0006 & 0.1794$\pm$0.0012 & \textbf{0.1513$\pm$0.0019} & 0.1828$\pm$0.0002 \\
& 336 & 0.2342 & 0.2365$\pm$0.0008 & 0.1880$\pm$0.0014 & \textbf{0.1664$\pm$0.0025} & 0.2339$\pm$0.0005 \\
& 720 & 0.3088 & 0.3165$\pm$0.0022 & 0.2141$\pm$0.0011 & \textbf{0.1855$\pm$0.0047} & 0.3011$\pm$0.0007 \\
\midrule
\multirow{4}{*}{Exchange}
& 96  & 0.0954 & 0.0922$\pm$0.0008 & 0.0892$\pm$0.0009 & \textbf{0.0783$\pm$0.0024} & 0.0948$\pm$0.0002 \\
& 192 & 0.1982 & 0.1904$\pm$0.0008 & 0.1095$\pm$0.0007 & \textbf{0.0949$\pm$0.0031} & 0.1934$\pm$0.0002 \\
& 336 & 0.3680 & 0.3218$\pm$0.0002 & 0.1132$\pm$0.0019 & \textbf{0.1102$\pm$0.0023} & 0.3601$\pm$0.0004 \\
& 720 & 0.9037 & 0.9291$\pm$0.0000 & 0.1935$\pm$0.0046 & \textbf{0.1859$\pm$0.0060} & 0.8866$\pm$0.0011 \\
\midrule
\multirow{4}{*}{Weather}
& 96  & 0.1506 & 0.1504$\pm$0.0009 & 0.1544$\pm$0.0001 & \textbf{0.1483$\pm$0.0019} & 0.1525$\pm$0.0001 \\
& 192 & 0.2011 & 0.2005$\pm$0.0007 & 0.1829$\pm$0.0003 & \textbf{0.1571$\pm$0.0017} & 0.1969$\pm$0.0002 \\
& 336 & 0.2566 & 0.2550$\pm$0.0006 & 0.2219$\pm$0.0009 & \textbf{0.1673$\pm$0.0036} & 0.2482$\pm$0.0004 \\
& 720 & 0.3384 & 0.3428$\pm$0.0017 & 0.2008$\pm$0.0001 & \textbf{0.1750$\pm$0.0028} & 0.3385$\pm$0.0021 \\
\bottomrule
\end{tabular}
\end{table*}

\paragraph{Implementation note.}
The official TimeXer implementation performs an in-place division during
input normalization.
This operation conflicts with repeated backpropagation through cached
inputs during TTA.
For TAFAS and PETSA, we replace it with the mathematically equivalent
non-in-place operation.
The numerical forward computation is unchanged.
COSA and \ours{} are unaffected by this implementation issue.

\subsection{Channel-Independent and Channel-Dependent Backbones}
\label{app:ci_cd}

We group the seven main backbones according to whether they explicitly
model cross-variate dependencies in their forecasting architecture.
DLinear, FreTS, OLS, and PatchTST are treated as channel-independent
(CI), while iTransformer, Informer, and MICN are treated as
channel-dependent (CD) under our experimental configurations.

\begin{table}[t]
  \centering
  \caption{
  Average MSE reduction of \ours{} over COSA by backbone family,
  averaged over six datasets and four horizons.
  }
  \label{tab:ci_cd}
  \vspace{2pt}
  \begin{tabular}{lcc}
  \toprule
  \textbf{Family} & \textbf{Backbones} &
  \textbf{MSE reduction vs.\ COSA} \\
  \midrule
  CI & DLinear, FreTS, OLS, PatchTST & 10.87\% \\
  CD & iTransformer, Informer, MICN & 10.18\% \\
  \bottomrule
  \end{tabular}
\end{table}

\begin{figure}[!t]
  \centering
  \parbox[t]{0.49\linewidth}{
    \centering
    \includegraphics[width=\linewidth]{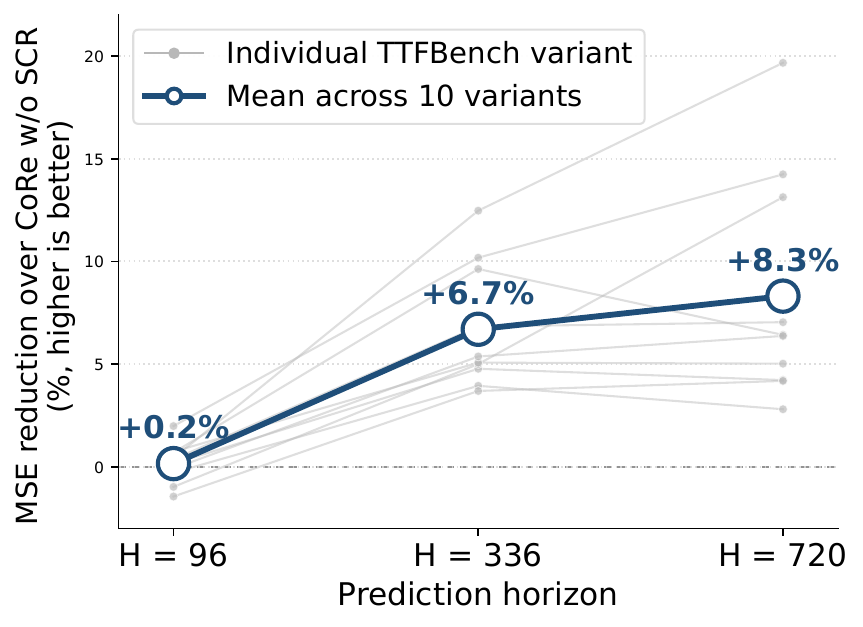}
    \captionof{figure}{
    Robustness on TTFBench (ETTh1) across ten distribution-shift variants.
    Grey curves denote individual variants and the solid curve denotes their mean.
    }
    \label{fig:ttfbench}
  }
  \hfill
  \parbox[t]{0.49\linewidth}{
    \centering
    \includegraphics[width=\linewidth]{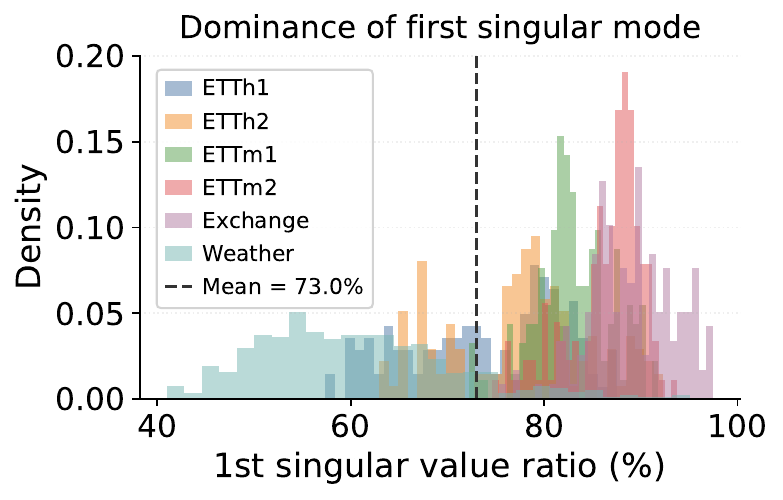}
    \captionof{figure}{
    Distribution of the leading singular-mode energy ratio of
    $\boldsymbol{\Delta}_t$.
    The first mode captures 73\% of correction energy on average.
    }
    \label{fig:singular_mode}
  }
\end{figure}

% Place Table~\ref{tab:rank_capacity_summary} right after Figures~\ref{fig:ttfbench} and~\ref{fig:singular_mode}
% so LaTeX can pack them onto the same float page if space allows.
\begin{figure}[t]
\centering
\small
\captionof{table}{Adapter-only parameter counts and accuracy for
three cross-variate interaction designs, averaged over
168 settings.}
\label{tab:rank_capacity_summary}
\vspace{2pt}
\begin{tabular}{lccc}
\toprule
& \textbf{Dense} & \textbf{SCR $r{=}C$}
& \textbf{SCR $r{=}16$} \\
\midrule
Avg params & 339,024 & 9,922 & 16,480 \\
Avg imp.\ vs.\ baseline & $+23.09\%$ & $+25.82\%$
& $+20.37\%$ \\
Win rate vs.\ baseline & 128/168 & 162/168 & 147/168 \\
\bottomrule
\end{tabular}
\end{figure}

\ours{} improves both groups by comparable margins, showing that
correction-space adaptation remains useful for both channel-independent
and channel-dependent forecasting backbones.
A summary is shown in Table~\ref{tab:ci_cd}.

\subsection{Robustness under Diverse Distribution Shifts}
\label{app:ttfbench}

TTFBench~\citep{grover2025shift} contains multiple types and intensities
of distribution shift for TSF-TTA evaluation.
Using ETTh1 with DLinear, we evaluate ten representative shift variants
with the same hyperparameter configuration as in the main experiments.

Figure~\ref{fig:ttfbench} reports the MSE reduction of \ours{} relative to
the corresponding variant without SCR.
The improvement remains positive across the evaluated shift types and is
generally larger at medium-to-long prediction horizons.

\section{Additional Ablation and Scalability Analysis}

\subsection{SCR Rank, Structure, and Scalability}
\label{app:rank_ablations}

Motivated by the low-dimensional correction structure observed in
Appendix~\ref{app:dataset_correlation}, we examine whether effective
cross-variate refinement requires dense interaction across all $C$
variates.

\paragraph{Correction-space structure.}
We first examine the singular-value spectrum of the learned correction
matrix $\boldsymbol{\Delta}_t\in\mathbb R^{H\times C}$.
Figure~\ref{fig:singular_mode} reports the energy captured by its dominant
singular mode.
Across settings, the leading mode captures approximately 73\% of the
correction energy on average, revealing pronounced low-dimensional
structure in the correction space.
This observation motivates the reduced-rank parameterizations evaluated
below.

% (Moved) Figure~\ref{fig:singular_mode} is placed together with Figure~\ref{fig:ttfbench} for a single-row layout.

\paragraph{Structured versus dense interaction.}
We compare dense mixing, fixed-rank SCR ($r=16$), and the
variate-matched configuration ($r=C$) over the 168 main settings.
Dense mixing uses approximately $34\times$ more parameters on average
than SCR with $r=C$, yet obtains a lower win rate
(128/168 versus 162/168) and a smaller average MSE reduction over the
frozen backbone (23.09\% versus 25.82\%).
Table~\ref{tab:rank_capacity_summary} summarizes the parameter cost and
aggregate performance, with full results reported in
Table~\ref{tab:core_rank_capacity_full}.

% Keep the rank/capacity tables together and avoid float spillover.
\FloatBarrier
\clearpage
%% ============================================================
%%  Rank/Capacity Ablation: CoRe (r=C) / CoRe (r=16) / CoRe (Dense)
%% ============================================================
\begin{sidewaystable*}[t]
\centering
\caption{
Full rank/capacity ablation results of \textsc{CoRe}. We compare the default variate-matched rank setting ($r{=}C$), a fixed-rank variant ($r{=}16$), and a dense horizon-space interaction without bottleneck.
The dense horizon-space interaction used in this comparison contains $2H^2+H$ parameters, whereas SCR with rank $r$ contains $3Hr+r+H$ parameters.
Lower MSE is better; \textbf{bold} and \underline{underlining} indicate the best and second-best results among the three variants, respectively.
}
\label{tab:core_rank_capacity_full}
\small
\setlength{\tabcolsep}{2.5pt}
\resizebox{\textwidth}{!}{%
\begin{tabular}{ll ccc ccc ccc ccc ccc ccc ccc}
\toprule
& & \multicolumn{3}{c}{\textbf{iTransformer}}
& \multicolumn{3}{c}{\textbf{DLinear}}
& \multicolumn{3}{c}{\textbf{FreTS}}
& \multicolumn{3}{c}{\textbf{OLS}}
& \multicolumn{3}{c}{\textbf{PatchTST}}
& \multicolumn{3}{c}{\textbf{MICN}}
& \multicolumn{3}{c}{\textbf{Informer}} \\
\cmidrule(lr){3-5}\cmidrule(lr){6-8}\cmidrule(lr){9-11}\cmidrule(lr){12-14}\cmidrule(lr){15-17}\cmidrule(lr){18-20}\cmidrule(lr){21-23}
& $H$
& $r{=}C$ & $r{=}16$ & \shortstack{Dense\\(horizon)}
& $r{=}C$ & $r{=}16$ & \shortstack{Dense\\(horizon)}
& $r{=}C$ & $r{=}16$ & \shortstack{Dense\\(horizon)}
& $r{=}C$ & $r{=}16$ & \shortstack{Dense\\(horizon)}
& $r{=}C$ & $r{=}16$ & \shortstack{Dense\\(horizon)}
& $r{=}C$ & $r{=}16$ & \shortstack{Dense\\(horizon)}
& $r{=}C$ & $r{=}16$ & \shortstack{Dense\\(horizon)} \\
\midrule

\multirow{4}{*}{\rotatebox[origin=c]{90}{\textbf{ETTh1}}}
& 96  & \underline{0.4545} & \textbf{0.4390} & 0.5133 & \underline{0.4916} & \textbf{0.4722} & 0.5133 & \underline{0.4625} & \textbf{0.4500} & 0.4638 & 0.4723 & \textbf{0.4517} & \underline{0.4707} & \underline{0.4371} & \textbf{0.4353} & 0.4396 & 0.5863 & \textbf{0.5461} & \underline{0.5839} & \underline{0.6501} & \textbf{0.6203} & 0.6543 \\
& 192 & \textbf{0.4589} & \underline{0.4767} & 0.4987 & \textbf{0.4961} & \underline{0.4963} & 0.4987 & \textbf{0.4692} & \underline{0.4806} & 0.4902 & \underline{0.4763} & 0.4786 & \textbf{0.4758} & \underline{0.4540} & 0.4685 & \textbf{0.4520} & \underline{0.5203} & 0.5290 & \textbf{0.5189} & 0.6415 & \textbf{0.6341} & \underline{0.6355} \\
& 336 & \textbf{0.4661} & 0.4918 & \underline{0.4686} & \underline{0.4722} & 0.5113 & \textbf{0.4686} & \textbf{0.4596} & 0.4979 & \underline{0.4686} & \underline{0.4568} & 0.4915 & \textbf{0.4558} & \underline{0.4458} & 0.4817 & \textbf{0.4413} & \underline{0.5386} & 0.5394 & \textbf{0.5380} & \underline{0.6302} & 0.6456 & \textbf{0.6290} \\
& 720 & \underline{0.5055} & 0.5387 & \textbf{0.4771} & \underline{0.4891} & 0.5330 & \textbf{0.4771} & \underline{0.5052} & 0.5482 & \textbf{0.4889} & \textbf{0.4886} & 0.5305 & \underline{0.4908} & \textbf{0.4983} & 0.5388 & \underline{0.5009} & \textbf{0.5591} & 0.5837 & \underline{0.5614} & \underline{0.6400} & 0.6524 & \textbf{0.6399} \\
\midrule

\multirow{4}{*}{\rotatebox[origin=c]{90}{\textbf{ETTh2}}}
& 96  & 0.2803 & \underline{0.2736} & \textbf{0.2460} & 0.2426 & \underline{0.2405} & \textbf{0.2383} & 0.2462 & \underline{0.2455} & \textbf{0.2422} & \underline{0.2389} & \textbf{0.2385} & 0.2434 & 0.2490 & \textbf{0.2437} & \underline{0.2447} & \textbf{0.2476} & 0.2551 & \underline{0.2497} & 0.3760 & \textbf{0.3476} & \underline{0.3571} \\
& 192 & \underline{0.2722} & 0.3154 & \textbf{0.2145} & \textbf{0.2275} & 0.2989 & \underline{0.2321} & \textbf{0.2371} & 0.2926 & \underline{0.2388} & \underline{0.2290} & 0.3004 & \textbf{0.2213} & \underline{0.2405} & 0.2590 & \textbf{0.2341} & \underline{0.2430} & 0.3096 & \textbf{0.2364} & \textbf{0.3567} & 0.3917 & \underline{0.3839} \\
& 336 & \underline{0.2245} & 0.2683 & \textbf{0.2031} & \underline{0.2171} & 0.2660 & \textbf{0.2158} & \underline{0.2196} & 0.2601 & \textbf{0.2189} & \textbf{0.2122} & 0.2641 & \underline{0.2130} & \underline{0.2167} & 0.2572 & \textbf{0.2164} & \underline{0.2486} & 0.2798 & \textbf{0.2478} & \textbf{0.3463} & 0.4037 & \underline{0.3498} \\
& 720 & \underline{0.2545} & 0.2744 & \textbf{0.2166} & \underline{0.2324} & 0.2449 & \textbf{0.2279} & \textbf{0.2336} & 0.2485 & \underline{0.2338} & \textbf{0.2301} & 0.2396 & \underline{0.2317} & \textbf{0.2331} & 0.2465 & \underline{0.2345} & \textbf{0.2746} & \underline{0.2829} & 0.2861 & \underline{0.3113} & 0.3405 & \textbf{0.3105} \\
\midrule

\multirow{4}{*}{\rotatebox[origin=c]{90}{\textbf{ETTm1}}}
& 96  & \underline{0.3776} & \textbf{0.3430} & 0.4121 & \underline{0.3881} & \textbf{0.3560} & 0.4121 & 0.3862 & \textbf{0.3584} & \underline{0.3776} & 0.3874 & \textbf{0.3556} & \underline{0.3831} & 0.4026 & \textbf{0.3656} & \underline{0.3979} & 0.4232 & \textbf{0.3724} & \underline{0.4219} & 0.4929 & \textbf{0.4130} & \underline{0.4876} \\
& 192 & \underline{0.4184} & \textbf{0.4052} & 0.4843 & 0.4282 & \textbf{0.4247} & \underline{0.4260} & \textbf{0.4200} & 0.4243 & \underline{0.4218} & \underline{0.4235} & \textbf{0.4230} & 0.4280 & \textbf{0.4183} & \underline{0.4215} & 0.4543 & 0.4551 & \textbf{0.4422} & \underline{0.4505} & 0.5385 & \textbf{0.4782} & \underline{0.5301} \\
& 336 & \underline{0.4585} & \textbf{0.4367} & 0.5230 & \underline{0.4665} & \textbf{0.4564} & 0.5230 & \underline{0.4616} & 0.4628 & \textbf{0.4575} & 0.4658 & \textbf{0.4626} & \underline{0.4640} & \textbf{0.4509} & \underline{0.4561} & 0.5043 & \underline{0.5018} & \textbf{0.4770} & 0.5026 & 0.5771 & \textbf{0.5270} & \underline{0.5766} \\
& 720 & \underline{0.5171} & \textbf{0.4708} & 0.5416 & 0.5424 & \textbf{0.4864} & \underline{0.5290} & \underline{0.5200} & \textbf{0.4683} & 0.5253 & \underline{0.5367} & \textbf{0.4783} & 0.5417 & \underline{0.5225} & \textbf{0.4692} & 0.5339 & 0.5602 & \textbf{0.4832} & \underline{0.5520} & 0.6454 & \textbf{0.5631} & \underline{0.6360} \\
\midrule

\multirow{4}{*}{\rotatebox[origin=c]{90}{\textbf{ETTm2}}}
& 96  & \underline{0.1831} & 0.1856 & \textbf{0.1623} & \underline{0.1644} & 0.1684 & \textbf{0.1623} & \underline{0.1707} & \textbf{0.1645} & 0.1710 & \underline{0.1655} & 0.1687 & \textbf{0.1650} & \underline{0.1721} & \textbf{0.1615} & 0.1732 & 0.1845 & \textbf{0.1687} & \underline{0.1723} & 0.2027 & \underline{0.2011} & \textbf{0.2001} \\
& 192 & \underline{0.1905} & 0.2074 & \textbf{0.1658} & \underline{0.1652} & 0.1904 & \textbf{0.1634} & \textbf{0.1628} & 0.1898 & \underline{0.1638} & \underline{0.1671} & 0.1930 & \textbf{0.1644} & \underline{0.1694} & 0.2065 & \textbf{0.1678} & \underline{0.1832} & 0.2035 & \textbf{0.1791} & \textbf{0.1964} & 0.2128 & \underline{0.1986} \\
& 336 & 0.2058 & \underline{0.2048} & \textbf{0.1720} & \underline{0.1725} & 0.1773 & \textbf{0.1720} & \textbf{0.1686} & 0.1761 & \underline{0.1690} & \textbf{0.1726} & 0.1758 & \underline{0.1742} & \underline{0.1765} & 0.1827 & \textbf{0.1752} & \underline{0.1871} & 0.1988 & \textbf{0.1836} & \underline{0.2114} & 0.2202 & \textbf{0.2105} \\
& 720 & \underline{0.1955} & 0.1964 & \textbf{0.1747} & 0.1778 & \underline{0.1776} & \textbf{0.1769} & \underline{0.1799} & 0.1803 & \textbf{0.1758} & \underline{0.1787} & \textbf{0.1773} & 0.1791 & \textbf{0.1728} & 0.1772 & \underline{0.1738} & \underline{0.1991} & \textbf{0.1973} & 0.2013 & \underline{0.2307} & 0.2345 & \textbf{0.2299} \\
\midrule

\multirow{4}{*}{\rotatebox[origin=c]{90}{\textbf{Exchange}}}
& 96  & \underline{0.0832} & 0.0916 & \textbf{0.0586} & \underline{0.0706} & 0.0886 & \textbf{0.0631} & \underline{0.0738} & 0.0815 & \textbf{0.0716} & \underline{0.0741} & 0.0808 & \textbf{0.0734} & \underline{0.0739} & 0.0800 & \textbf{0.0709} & \underline{0.0779} & 0.0963 & \textbf{0.0756} & \textbf{0.0985} & 0.1482 & \underline{0.1003} \\
& 192 & \underline{0.0988} & 0.1453 & \textbf{0.0728} & \underline{0.0733} & 0.1033 & \textbf{0.0728} & \textbf{0.0806} & 0.1116 & \underline{0.0848} & \textbf{0.0812} & 0.1106 & \underline{0.0819} & \underline{0.0897} & 0.1153 & \textbf{0.0869} & \textbf{0.0739} & 0.1078 & \underline{0.0757} & \textbf{0.1183} & 0.1568 & \underline{0.1203} \\
& 336 & \underline{0.1054} & 0.1419 & \textbf{0.0721} & \textbf{0.0804} & 0.1124 & \underline{0.0825} & \textbf{0.0933} & 0.1254 & \underline{0.0962} & \underline{0.0947} & 0.1247 & \textbf{0.0915} & \underline{0.1000} & 0.1323 & \textbf{0.0987} & \underline{0.1091} & 0.1528 & \textbf{0.1083} & \underline{0.1901} & 0.2537 & \textbf{0.1880} \\
& 720 & \underline{0.1957} & 0.2597 & \textbf{0.1310} & \textbf{0.1239} & 0.1790 & \underline{0.1244} & \textbf{0.1289} & 0.1921 & \underline{0.1371} & \textbf{0.1259} & 0.1833 & \underline{0.1313} & \textbf{0.1348} & 0.1930 & \underline{0.1375} & \underline{0.2211} & 0.3579 & \textbf{0.2194} & \textbf{0.3403} & 0.6156 & \underline{0.3652} \\
\midrule

\multirow{4}{*}{\rotatebox[origin=c]{90}{\textbf{Weather}}}
& 96  & \underline{0.1635} & \textbf{0.1588} & 0.3015 & \textbf{0.1719} & \underline{0.1732} & 0.1801 & \textbf{0.1617} & 0.1703 & \underline{0.1632} & \textbf{0.1704} & \underline{0.1723} & 0.1754 & \underline{0.1622} & 0.1644 & \textbf{0.1602} & \underline{0.1646} & \textbf{0.1636} & 0.1652 & \underline{0.1807} & \textbf{0.1751} & 0.1815 \\
& 192 & \textbf{0.1697} & 0.1936 & \underline{0.1823} & \textbf{0.1711} & 0.2012 & \underline{0.1726} & \underline{0.1690} & 0.2030 & \textbf{0.1689} & \textbf{0.1719} & 0.2018 & \underline{0.1777} & \underline{0.1724} & 0.2002 & \textbf{0.1675} & \textbf{0.1608} & \underline{0.1919} & \textbf{0.1608} & \textbf{0.2261} & 0.2336 & \underline{0.2316} \\
& 336 & \textbf{0.1736} & 0.2050 & \underline{0.1907} & \underline{0.1782} & 0.1992 & \textbf{0.1780} & \textbf{0.1785} & 0.1998 & \underline{0.1801} & \textbf{0.1777} & 0.1988 & \underline{0.1796} & \underline{0.1810} & 0.2045 & \textbf{0.1760} & \underline{0.1698} & 0.2072 & \textbf{0.1689} & \textbf{0.2202} & 0.2527 & \underline{0.2205} \\
& 720 & \textbf{0.1876} & 0.2157 & \underline{0.1994} & \textbf{0.1777} & 0.2127 & \underline{0.1792} & \textbf{0.1763} & 0.2114 & \underline{0.1769} & \textbf{0.1785} & 0.2136 & \underline{0.1810} & \underline{0.1765} & 0.2037 & \textbf{0.1754} & \textbf{0.1711} & 0.1999 & \underline{0.1712} & \textbf{0.2371} & 0.2439 & \underline{0.2399} \\
\bottomrule
\end{tabular}}
\end{sidewaystable*}
\clearpage
\FloatBarrier

\paragraph{Parameter scaling.}
The SCR bottleneck uses
\[
\mathbf W_{\downarrow}\in\mathbb R^{r\times 2H},
\qquad
\mathbf W_{\uparrow}\in\mathbb R^{H\times r}.
\]
Ignoring bias terms, SCR therefore contains
\[
2Hr+Hr=3Hr
\]
weights. With the default $r=C$, its parameter count scales as $O(HC)$;
with a fixed $r\ll C$, it scales as $O(Hr)$.

For comparison, the dense interaction used in the rank/capacity
ablation is implemented as a single \texttt{Linear}($2H\!\to\!H$)
layer (Table~\ref{tab:rank_capacity_summary}), containing $2H^2$ weights
(and $H$ bias parameters).
\paragraph{Rank sensitivity and default choice.}
We additionally compare smaller fixed ranks
$r\in\{4,8,16,32\}$ with the default $r=C$.
Among the evaluated configurations on the six standard datasets
(Table~\ref{tab:pool_rank}),
$r=C$ provides the strongest aggregate performance.
Smaller ranks remain effective in many settings and are useful when
reducing the parameter cost is important.
We therefore use $r=C$ as the default for the main experiments and
$r=16\ll C$ for the high-dimensional experiments.

\begin{table}[H]
\centering
\caption{Ablation on SCR pooling strategy and bottleneck rank (DLinear backbone, 24 settings). \emph{avg\_imp}: average MSE improvement over backbone.}
\label{tab:pool_rank}
\small
\begin{tabular}{lc}
\toprule
\textbf{Configuration} & \textbf{avg\_imp} \\
\midrule
Mean pooling, $r=4$  & +18.09\% \\
Mean pooling, $r=8$  & +18.84\% \\
Mean pooling, $r=16$ & +19.66\% \\
Mean pooling, $r=32$ & +20.24\% \\
Mean pooling, $r=C$  & \textbf{+25.45\%} \\
\midrule
Max pooling, $r=16$  & +19.36\% \\
\bottomrule
\end{tabular}
\end{table}

\paragraph{High-Dimensional Datasets.}\label{app:highdim}

The default SCR configuration uses $r=C$, so its parameter cost increases
with the number of variates.
We therefore examine whether correction-space interaction remains useful
with a substantially smaller bottleneck rank in high-dimensional
settings.

Table~\ref{tab:highdim} reports results on Electricity ($C=321$) and
Traffic ($C=862$) using $r=16\ll C$.
Despite the substantial rank reduction, \textsc{CoRe} retains positive
aggregate improvements over COSA.
The gains are particularly clear on Electricity.
On Traffic, the improvements are smaller and several individual settings
show negative differences, including DLinear at $H=720$.

Together with the singular-value analysis in
Appendix~\ref{app:rank_ablations}, these results show that useful
cross-variate correction structure can be retained with a bottleneck
substantially smaller than $C$ in high-dimensional settings.
These results show that aggregate improvements can be retained with a
bottleneck rank substantially smaller than $C$, while also highlighting
the limits of a single fixed rank at very large $C$.

\begin{table}[t]
\centering
\caption{Results on large-scale datasets Electricity 
($C{=}321$) and Traffic ($C{=}862$), averaged over 
10 seeds.  
\textbf{COSA$\to$\ours{}}: MSE improvement of 
\textsc{CoRe} over COSA. All methods use a fixed 
bottleneck rank $r{=}16$ for these datasets to keep the parameter cost independent of $C$.}
\label{tab:highdim}
\small
\setlength{\tabcolsep}{4pt}
\begin{tabular}{llrcccc}
\toprule
\textbf{Model} & \textbf{Dataset} & $H$ &
\textbf{Backbone} & \textbf{COSA} & 
\textbf{\textsc{CoRe}} &
\textbf{COSA$\to$\ours{}} \\
\midrule
\multirow{8}{*}{DLinear}
& \multirow{4}{*}{Electricity}
& 96  & 0.2078 & 0.2024 & \textbf{0.1992} & $+1.58\%$ \\
&& 192 & 0.2081 & 0.1922 & \textbf{0.1771} & $+7.86\%$ \\
&& 336 & 0.2228 & 0.1947 & \textbf{0.1751} & $+10.07\%$ \\
&& 720 & 0.2644 & 0.2139 & \textbf{0.1926} & $+9.96\%$ \\
\cmidrule(lr){2-7}
& \multirow{4}{*}{Traffic}
& 96  & 0.6710 & 0.6685 & \textbf{0.6648} & $+0.55\%$ \\
&& 192 & 0.6251 & 0.6127 & \textbf{0.6031} & $+1.57\%$ \\
&& 336 & 0.6328 & 0.6133 & \textbf{0.6074} & $+0.96\%$ \\
&& 720 & 0.6717 & \textbf{0.6464} & 0.6545 & $-1.25\%$ \\
\midrule
\multirow{8}{*}{MICN}
& \multirow{4}{*}{Electricity}
& 96  & 0.1766 & 0.1634 & \textbf{0.1526} & $+6.61\%$ \\
&& 192 & 0.1915 & 0.1592 & \textbf{0.1409} & $+11.49\%$ \\
&& 336 & 0.1973 & 0.1575 & \textbf{0.1396} & $+11.37\%$ \\
&& 720 & 0.2193 & 0.1674 & \textbf{0.1501} & $+10.33\%$ \\
\cmidrule(lr){2-7}
& \multirow{4}{*}{Traffic}
& 96  & 0.4955 & \textbf{0.4602} & 0.4603 & $-0.02\%$ \\
&& 192 & 0.5080 & 0.4737 & \textbf{0.4449} & $+6.08\%$ \\
&& 336 & 0.5306 & 0.4929 & \textbf{0.4642} & $+5.82\%$ \\
&& 720 & 0.5757 & 0.5232 & \textbf{0.5083} & $+2.85\%$ \\
\bottomrule
\end{tabular}
\end{table}
 
\subsection{Spectral Descriptor and Gate Ablations}
\label{app:spectral_full_ablations}

Table~\ref{tab:ablation} evaluates alternative spectral descriptors and
gating variants.
Differences among the spectral-descriptor variants are small when
averaged across settings.
We use the full four-dimensional descriptor
\[
[
\mathrm{SE},
\mathrm{LBR},
\mathrm{MBR},
\mathrm{HBR}
]
\]
as the default because it jointly summarizes spectral concentration and
coarse frequency allocation at negligible additional cost.

The fixed-gate and input-conditioned variants further isolate the
contribution of adaptive modulation.
The full spectral gate provides a smaller additional average gain beyond
the structured SCR interaction.

\begin{table}[!ht] 
\centering
\caption{Ablation study of \textsc{CoRe} components.
Average MSE over three backbones (DLinear, PatchTST, MICN),
six datasets, four horizons, and 10 seeds.
\textbf{vs.\ COSA}: relative MSE reduction over COSA.
\emph{Loss-trend gate}: COSA's adaptive learning-rate
signal repurposed as the SCR gate.
\emph{Fixed gate} ($g{=}1$): uniform cross-variate
refinement without input-dependent modulation.
$\mathrm{SE}_t$: spectral entropy; LBR/MBR/HBR$_t$: low/mid/high
band energy ratios (Section~\ref{sec:spectral}).}
\label{tab:ablation}
\setlength{\tabcolsep}{2pt}
\renewcommand{\arraystretch}{0.85}
\resizebox{\linewidth}{!}{%
\begin{tabular}{lccccc}
\toprule
\textbf{Method} & 
\textbf{DLinear} & \textbf{PatchTST} & \textbf{MICN} &
\textbf{Avg.\ MSE} & \textbf{vs.\ COSA} \\
\midrule
COSA & 0.2981 & 0.2927 & 0.3241 & 0.3050 & -- \\
\midrule
+SCR (no anchor; per-variate bottleneck) &
0.2760 & 0.2721 & 0.3018 & 0.2833 & $-7.1\%$ \\
+SCR (no anchor/bottleneck; full $C{\times}C$ mixing) &
0.2745 & 0.2706 & 0.3044 & 0.2832 & $-7.2\%$ \\
\midrule
+SCR (loss-trend gate) & 
0.2712 & 0.2657 & 0.2963 & 0.2777 & $-9.0\%$ \\
+SCR (fixed gate, $g{=}1$) & 
0.2703 & 0.2651 & 0.2954 & 0.2769 & $-9.2\%$ \\
\midrule
+SCR ($\mathbf{s}_t{=}[\mathrm{SE}_t]$) & 
0.2676 & 0.2638 & 0.2946 & 0.2753 & $-9.7\%$ \\
+SCR ($\mathbf{s}_t{=}[\mathrm{SE}_t, \mathrm{LBR}_t]$) & 
0.2677 & 0.2638 & 0.2938 & 0.2751 & $-9.8\%$ \\
+SCR ($\mathbf{s}_t{=}[\mathrm{SE}_t, \mathrm{LBR}_t, 
\mathrm{MBR}_t]$) & 
0.2680 & 0.2641 & 0.2941 & 0.2754 & $-9.7\%$ \\
\midrule
\textsc{CoRe}-PV ($\mathbf{s}_t^{(c)} {=}[\mathrm{SE}_t^{(c)}, \mathrm{LBR}_t^{(c)}, 
\mathrm{MBR}_t^{(c)}, \mathrm{HBR}_t^{(c)}]$) &
0.2671 & 0.2640 & 0.2945 & 0.2752 & $-9.8\%$ \\
\textsc{CoRe} ($\mathbf{s}_t{=}[\mathrm{SE}_t, \mathrm{LBR}_t, 
\mathrm{MBR}_t, \mathrm{HBR}_t]$) & 
0.2675 & 0.2646 & 0.2942 & 
0.2754 & $-9.7\%$ \\
\bottomrule
\end{tabular}%
}
\end{table}

\subsection{Spectral Gate Analysis}
\label{app:gating}

\paragraph{Shared versus per-variate descriptors.}
The main implementation computes a shared spectral descriptor by averaging
the power spectra across variates before constructing
$\mathbf s_t$.
We additionally evaluate per-variate spectral descriptors.
The two variants perform comparably
(average MSE 0.2752 versus 0.2754), so we retain the shared descriptor as
the simpler default.

\begin{figure}[t]
  \centering
  \includegraphics[width=1.0\linewidth]
  {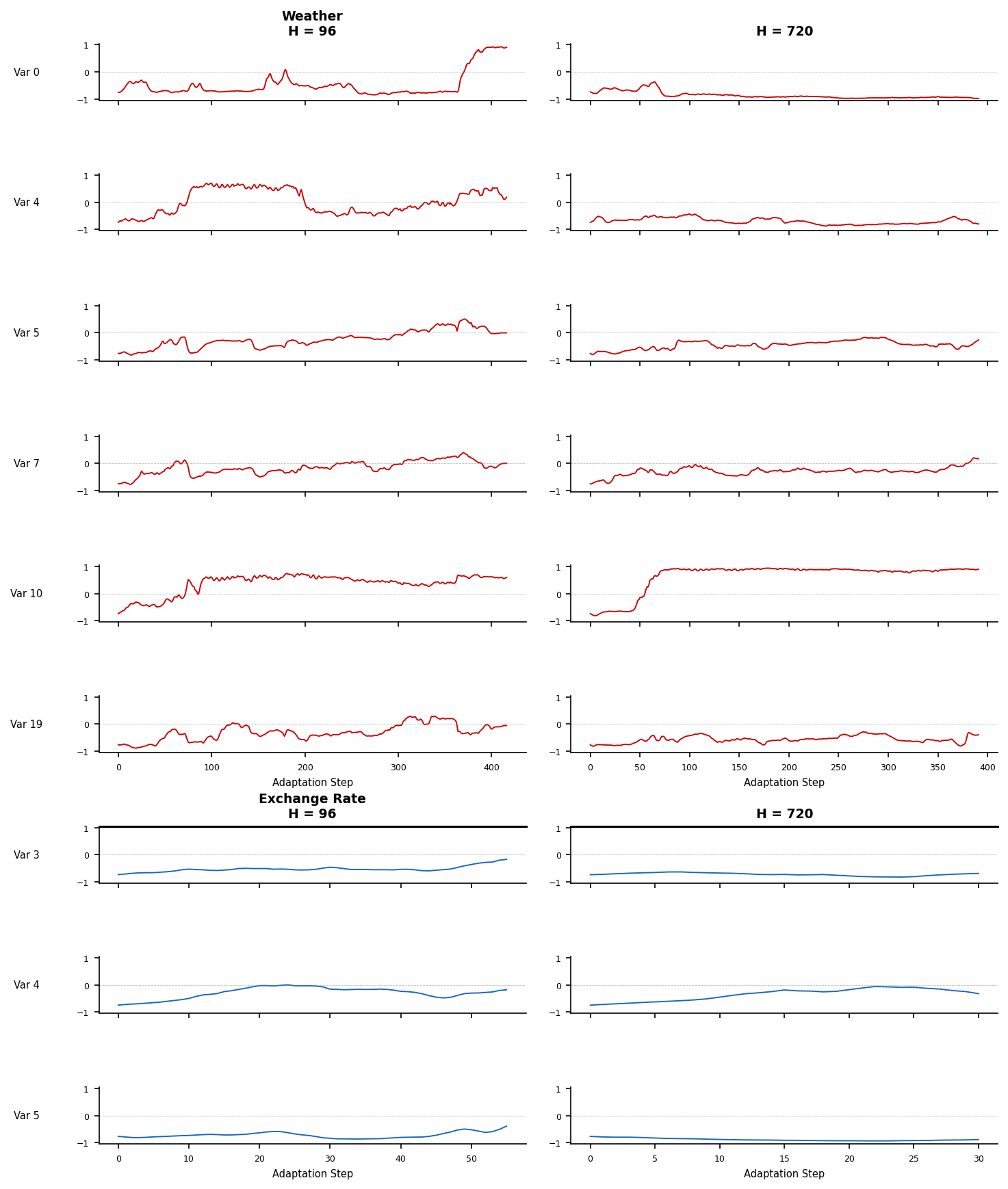}
  \caption{
  Example spectral-gate trajectories over adaptation steps on Weather
  and Exchange Rate.
  }
  \label{fig:gate_vis_combined}
\end{figure}

\paragraph{Gate behavior.}
Table~\ref{tab:gate_stats} summarizes statistics of the learned gates,
and Figure~\ref{fig:gate_vis_combined} shows representative gate
trajectories on Weather and Exchange Rate.
The gate varies substantially across datasets and test batches, with
approximately 10.5\% of activations near zero.
This shows that the spectral controller produces input-dependent modulation rather than applying SCR with a fixed contribution.
Together with the fixed-gate ablation in Appendix~\ref{app:spectral_full_ablations}, these statistics show that the gate is input-dependent and provides an empirical benefit over fixed modulation.

\begin{table}[H]
\centering
\caption{SCR spectral gate statistics across datasets
(DLinear backbone, averaged over 4 horizons and 10
seeds). Mean: time-averaged gate value across
all variates and test batches; Std: standard
deviation across variates and batches;
Near zero ($|g|{<}0.1$): fraction of gate
activations close to zero, indicating suppressed
cross-variate refinement;
Near init ($g{<}{-}0.8$): fraction of gate
activations close to the initialisation bias
$\mathbf{b}_g{=}{-1}$, indicating gates that have
not moved significantly from initialisation.
The gate is bounded in $(-1, 1)$ by $\tanh$.}
\label{tab:gate_stats}
\small
\setlength{\tabcolsep}{6pt}
\begin{tabular}{lcccc}
\toprule
\textbf{Dataset} & 
\textbf{Mean} & 
\textbf{Std} & 
\textbf{Near zero} ($|g|{<}0.1$) & 
\textbf{Near init} ($g{<}{-}0.8$) \\
\midrule
ETTh1        & $-0.583$ & $0.325$ & $5.9\%$  & $34.5\%$ \\
ETTh2        & $-0.631$ & $0.285$ & $9.3\%$  & $37.5\%$ \\
ETTm1        & $-0.357$ & $0.421$ & $9.9\%$  & $15.1\%$ \\
ETTm2        & $-0.498$ & $0.354$ & $15.5\%$ & $28.4\%$ \\
Exchange     & $-0.676$ & $0.194$ & $2.4\%$  & $23.9\%$ \\
Weather      & $-0.352$ & $0.371$ & $20.3\%$ & $9.0\%$  \\
\midrule
Average      & $-0.516$ & $0.325$ & $10.5\%$ & $24.7\%$ \\
\bottomrule
\end{tabular}
\end{table}

\paragraph{Gate-bias sensitivity.}
Table~\ref{tab:sensitivity} evaluates different spectral-gate bias
initializations over the reported DLinear settings.
The default
$\mathbf b_g=-\mathbf1$
achieves the strongest aggregate performance.
Relative to this choice, initialization at
$\mathbf b_g=\mathbf0$
increases MSE by 1.56\%.

\begin{table}[H]
\centering
\caption{Gate bias sensitivity.}
\label{tab:sensitivity}
\small
\begin{tabular}{cccc}
\toprule
$\mathbf{b}_g$ & $-2.0$ & $-1.0$ & $0.0$ \\
\midrule
vs.\ default $\downarrow$ & $+0.46\%$ & $+0.00\%$ & $+1.56\%$ \\
\bottomrule
\end{tabular}
\end{table}

\section{Online Adaptation Reliability}
\label{app:eval_framework}

We evaluate adaptation reliability relative to the frozen backbone using
the mean per-batch regret
\begin{equation}
    \overline{\mathcal R}_T(m)
    =
    \frac{1}{T}
    \sum_{t=1}^{T}
    \left(
        \ell_t^{m}
        -
        \ell_t^{\mathrm{frozen}}
    \right),
\end{equation}
where $m$ denotes the adapted method and negative values indicate
an average improvement over the frozen backbone.

We additionally report three complementary quantities.
The positive-part regret
$\overline{\mathcal R}_T^+$
measures the magnitude of harmful adaptation when it occurs;
$p_{\mathrm{neg}}$ denotes the frequency with which the adapted model
underperforms the frozen backbone; and $n_{\mathrm{worse}}$ counts the 18 horizon-averaged backbone--dataset combinations with positive mean regret. Together, these metrics characterize two complementary aspects of reliability: how often online adaptation is harmful and how severe the harm is when it occurs.

Table~\ref{tab:backbone_consistency} reports the full reliability results
across the 18 backbone--dataset combinations.
\begin{table*}[t]
\centering
\caption{Online adaptation consistency over 18 backbone--dataset combinations (3 backbones $\times$ 6 datasets), each averaged over $H\in\{96,192,336,720\}$.
$\overline{\mathcal{R}}_T$: mean per-batch
regret relative to the frozen backbone
(negative = better than backbone on average);
$p_{\text{neg}}$: fraction of batches where
the adapted model underperforms the backbone.
Best per row in \textbf{bold}. Overall row
also reports $\overline{\mathcal{R}}_T^{+}$
(average regret on settings where the method
is worse than the backbone) and
$n_{\text{worse}}$: number of backbone--dataset combinations
(out of 18) with positive mean regret.}
\label{tab:backbone_consistency}
\footnotesize
\setlength{\tabcolsep}{2pt}
% More vertical breathing room (the previous 0.85 was quite tight)
\renewcommand{\arraystretch}{0.8}
% Let booktabs use its default rule spacing (less cramped than manual overrides)
%\setlength{\aboverulesep}{0.4pt}
%\setlength{\belowrulesep}{0.4pt}
%\setlength{\cmidrulesep}{0.2pt}
\begin{tabular}{llcccccc}
\toprule
& &
\multicolumn{2}{c}{\textbf{TAFAS}} &
\multicolumn{2}{c}{\textbf{COSA}} &
\multicolumn{2}{c}{\textbf{\textsc{CoRe}}} \\
\cmidrule(lr){3-4}\cmidrule(lr){5-6}\cmidrule(lr){7-8}
\textbf{Model} & \textbf{Dataset} &
$\overline{\mathcal{R}}_T$ & $p_{\text{neg}}$ &
$\overline{\mathcal{R}}_T$ & $p_{\text{neg}}$ &
$\overline{\mathcal{R}}_T$ & $p_{\text{neg}}$ \\
\midrule
\multirow{6}{*}{DLinear}
& ETTh1
& $-0.003$ & $0.519$
& $-0.041$ & $0.402$
& $\mathbf{-0.080}$ & $\mathbf{0.300}$ \\
& ETTh2
& $+0.003$ & $0.570$
& $-0.056$ & $0.279$
& $\mathbf{-0.089}$ & $\mathbf{0.223}$ \\
& ETTm1
& $-0.024$ & $\mathbf{0.330}$
& $+0.013$ & $0.383$
& $\mathbf{-0.035}$ & $0.382$ \\
& ETTm2
& $+0.001$ & $0.546$
& $-0.026$ & $0.391$
& $\mathbf{-0.055}$ & $\mathbf{0.330}$ \\
& Exchange
& $-0.013$ & $0.213$
& $-0.262$ & $0.142$
& $\mathbf{-0.274}$ & $\mathbf{0.088}$ \\
& Weather
& $-0.011$ & $0.395$
& $-0.061$ & $0.183$
& $\mathbf{-0.094}$ & $\mathbf{0.139}$ \\
\midrule
\multirow{6}{*}{MICN}
& ETTh1
& $-0.030$ & $0.419$
& $-0.145$ & $0.190$
& $\mathbf{-0.187}$ & $\mathbf{0.146}$ \\
& ETTh2
& $+0.011$ & $0.531$
& $-0.066$ & $0.244$
& $\mathbf{-0.098}$ & $\mathbf{0.189}$ \\
& ETTm1
& $-0.020$ & $0.413$
& $-0.013$ & $0.335$
& $\mathbf{-0.043}$ & $\mathbf{0.331}$ \\
& ETTm2
& $+0.004$ & $0.549$
& $-0.025$ & $0.387$
& $\mathbf{-0.060}$ & $\mathbf{0.308}$ \\
& Exchange
& $-0.127$ & $0.364$
& $-0.403$ & $0.128$
& $\mathbf{-0.419}$ & $\mathbf{0.095}$ \\
& Weather
& $+0.037$ & $0.629$
& $-0.059$ & $0.194$
& $\mathbf{-0.089}$ & $\mathbf{0.159}$ \\
\midrule
\multirow{6}{*}{PatchTST}
& ETTh1
& $-0.007$ & $0.434$
& $-0.057$ & $0.349$
& $\mathbf{-0.089}$ & $\mathbf{0.268}$ \\
& ETTh2
& $+0.000$ & $0.444$
& $-0.057$ & $0.280$
& $\mathbf{-0.076}$ & $\mathbf{0.245}$ \\
& ETTm1
& $-0.010$ & $0.348$
& $-0.001$ & $0.350$
& $\mathbf{-0.026}$ & $\mathbf{0.352}$ \\
& ETTm2
& $+0.004$ & $0.522$
& $-0.025$ & $0.376$
& $\mathbf{-0.059}$ & $\mathbf{{0.307}}$ \\
& Exchange
& $-0.003$ & $0.389$
& $-0.243$ & $0.161$
& $\mathbf{-0.261}$ & $\mathbf{0.131}$ \\
& Weather
& $+0.005$ & $0.417$
& $-0.041$ & $0.207$
& $\mathbf{-0.082}$ & $\mathbf{0.173}$ \\
\midrule
\multicolumn{2}{l}{\textbf{Overall avg}}
& $-0.010$ & $0.446$
& $-0.087$ & $0.277$
& $\mathbf{-0.117}$ & $\mathbf{0.232}$ \\
\multicolumn{2}{l}{$\overline{\mathcal{R}}_T^{+}$}
& $0.0138$ & & $0.0127$ & & $\mathbf{0.0108}$ & \\
\multicolumn{2}{l}{$n_{\text{worse}}$ / 18}
& $8$ & & $5$ & & $\mathbf{2}$ & \\
\bottomrule
\end{tabular}
\end{table*}

\section{Additional Diagnostics}
\label{app:diagnostics}

\subsection{Dataset and Correction Statistics}
\label{app:dataset_correlation}

\paragraph{Cross-variate correlations.}
Table~\ref{tab:correlation} reports descriptive statistics of
cross-variate correlations across the evaluated datasets.
Before computing static correlations, we apply the Augmented
Dickey--Fuller test to each variate.
Differencing is applied when the unit-root null cannot be rejected at the
chosen threshold: order 1 for Exchange Rate and order 2 for Electricity.
Temporal correlation statistics are computed on the raw series.

The six main datasets have broadly comparable mean static absolute
correlations, while their temporal variation differs substantially.
Exchange Rate has the largest shifting standard deviation among the
reported datasets
(std $=0.479$),
whereas Traffic and Electricity exhibit more stable correlation
statistics.
These measurements provide descriptive context for the diversity of
cross-variate structure represented by the benchmark suite.

\begin{table}[tbp]
\centering
\caption{Pairwise Pearson correlation statistics across variates.
$N$: number of variates; Window: rolling window size (approx.\ 1 week
in physical time). \textbf{Mean $|r|$}: time-averaged absolute pairwise
correlation computed on stationary series after ADF test.
\textbf{Shifting (std)}: standard deviation of windowed correlations
over time, measuring temporal instability of cross-variate dependencies
(computed on raw series to preserve trend-switching signals).
\textbf{CV}: coefficient of variation (std/mean).
\textbf{Range}: peak-to-trough variation of windowed correlations.
\textbf{Strong} ($|r|{>}0.7$) and \textbf{Weak} ($|r|{\leq}0.3$):
fraction of variate pairs by static $|r|$ strength.
Datasets above the divider are used in main experiments.}
\label{tab:correlation}
\footnotesize
\setlength{\tabcolsep}{3pt}
\renewcommand{\arraystretch}{0.95}
% \resizebox{\linewidth}{!}{...} will always scale back up to full width, canceling font-size changes.
\scalebox{0.93}{%
\begin{tabular}{lrrrrrrrr}
\toprule
\textbf{Dataset} & $N$ & \textbf{Window} &
\textbf{Mean $|r|$} &
\textbf{Shifting (std)} &
\textbf{CV} &
\textbf{Range} &
\textbf{Strong} &
\textbf{Weak} \\
\midrule
ETTh1         &   7 & 168  & 0.222 & 0.266 & 0.919 & 0.771 &  9.5\% & 81.0\% \\
ETTh2         &   7 & 168  & 0.325 & 0.237 & 0.846 & 0.865 &  4.8\% & 57.1\% \\
ETTm1         &   7 & 672  & 0.224 & 0.264 & 0.913 & 0.771 &  9.5\% & 81.0\% \\
ETTm2         &   7 & 672  & 0.325 & 0.234 & 0.845 & 0.872 &  4.8\% & 57.1\% \\
Weather       &  21 & 1008 & 0.296 & 0.180 & 0.736 & 0.637 & 21.0\% & 64.3\% \\
Exchange Rate$^\dagger$ &   8 &  30  & 0.305 & \textbf{0.479} & \textbf{0.957} & \textbf{0.992} &  3.6\% & 42.9\% \\
\midrule
Traffic       & 862 & 168  & 0.631 & 0.111 & 0.207 & 0.651 & 25.7\% & 10.1\% \\
Electricity$^{\dagger\dagger}$ & 321 & 168 & 0.119 & 0.114 & 0.340 & 0.524 &  0.0\% & 91.0\% \\
\bottomrule
\multicolumn{9}{l}{$^\dagger$ Static Mean $|r|$ computed after first-order differencing (ADF $p < 0.05$);} \\
\multicolumn{9}{l}{\phantom{$^\dagger$} Shifting metrics computed on raw series to preserve trend-switching signals.} \\
\multicolumn{9}{l}{$^{\dagger\dagger}$ Second-order differencing applied (ADF $p < 0.05$).}
\end{tabular}%
}
\end{table}

\paragraph{Correction heterogeneity.}
Table~\ref{tab:gate_motivation} additionally reports variation among
per-variate correction signals.
The coefficient of variation and cosine-similarity statistics show that
the learned corrections can differ across variates in both magnitude and
direction.
We use these measurements as descriptive properties of the correction
space rather than as causal explanations of the gating results.

\begin{table}[tbp]
\centering
\caption{Per-variate correction heterogeneity 
and the benefit of spectral gating over fixed 
gating (DLinear backbone). 
\emph{CV}: coefficient of variation of 
per-variate correction norms across variates; 
\emph{Cos-sim std}: standard deviation of 
cosine similarity between each variate's 
correction and the cross-variate mean anchor; 
\emph{Fixed gate} and \emph{Spectral gate}: 
average MSE under each gating strategy 
(10 seeds); \emph{Gain}: relative improvement 
from spectral over fixed gating. 
Datasets ordered by CV.}
\label{tab:gate_motivation}
\footnotesize
\setlength{\tabcolsep}{5pt}
\begin{tabular}{lccccc}
\toprule
\textbf{Dataset} & 
\textbf{CV} & 
\textbf{Cos-sim std} &
\textbf{Fixed gate} & 
\textbf{Spectral gate} &
\textbf{Gain} \\
\midrule
ETTh1    & $0.217$ & $0.097$ & $0.4889$ & $\mathbf{0.4873}$ & $+0.34\%$ \\
Exchange & $0.322$ & $0.223$ & $0.0890$ & $\mathbf{0.0871}$ & $+2.19\%$ \\
ETTm1    & $0.428$ & $0.117$ & $0.4598$ & $\mathbf{0.4563}$ & $+0.76\%$ \\
ETTm2    & $0.514$ & $0.137$ & $0.1716$ & $\mathbf{0.1700}$ & $+0.95\%$ \\
ETTh2    & $0.554$ & $0.119$ & $0.2329$ & $\mathbf{0.2299}$ & $+1.29\%$ \\
Weather  & $0.696$ & $0.151$ & $0.1797$ & $\mathbf{0.1747}$ & $+2.77\%$ \\
\midrule
Average  & $0.455$ & $0.141$ & $0.2703$ & $\mathbf{0.2675}$ & $+1.03\%$ \\
\bottomrule
\end{tabular}
\end{table}

\paragraph{Low-dimensional correction structure.}
Across settings, the leading singular mode captures approximately 73\% of
the correction energy on average, revealing pronounced low-dimensional
structure in the correction space.
This observation motivates the reduced-rank parameterization examined in
Appendix~\ref{app:rank_ablations}.

\subsection{Qualitative Forecast Visualization}
\label{app:qual_exchange_vis}

\begin{figure*}[t]
    \centering
    \includegraphics[width=0.95\textwidth,height=0.85\textheight,keepaspectratio]{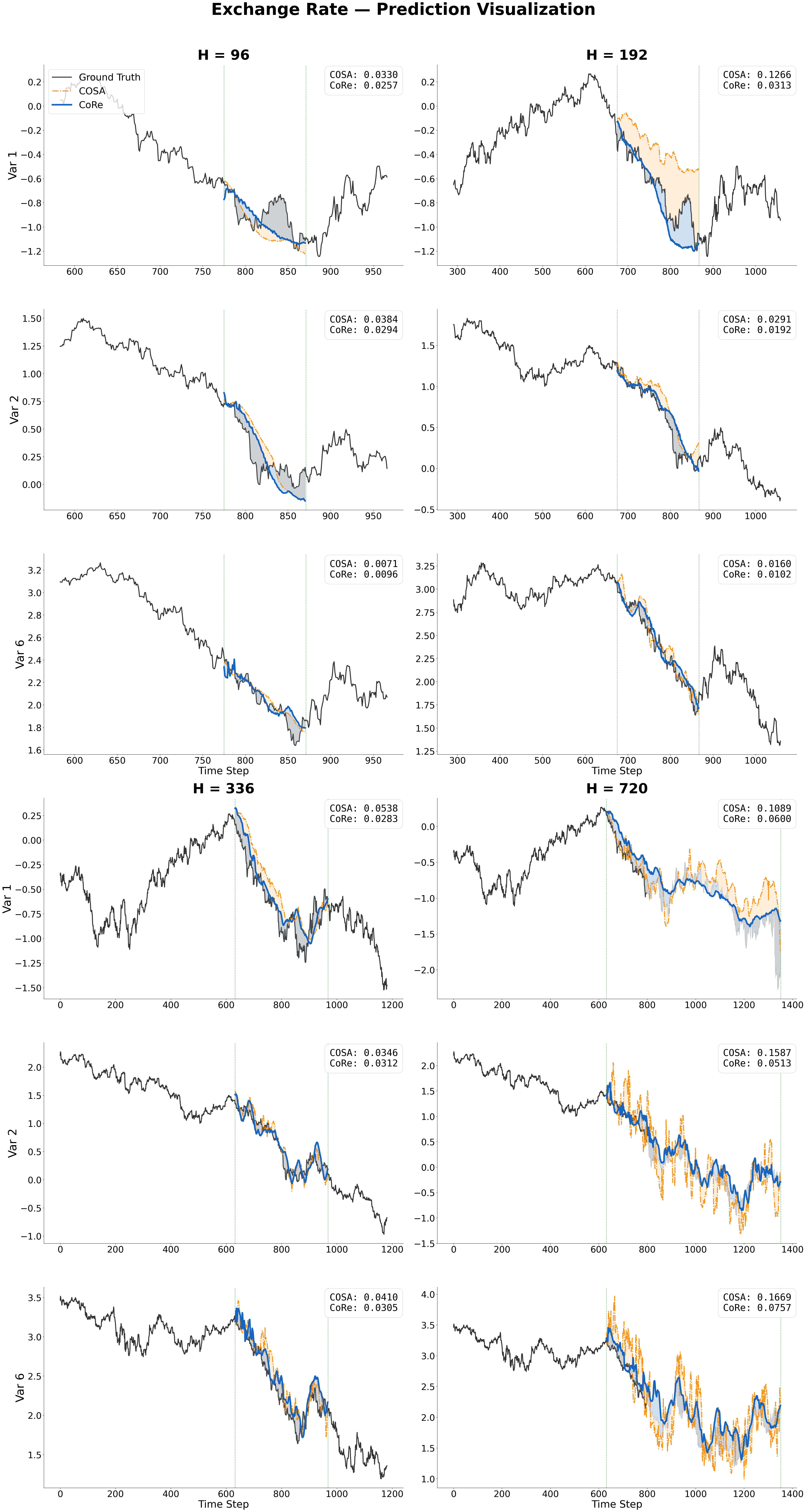}
    \caption{
    Qualitative comparison between COSA and \textsc{CoRe} on Exchange Rate
    across four prediction horizons and three selected variates.
    }
    \label{fig:pred_vis_exchange_rate}
\end{figure*}

\begin{figure*}[t]
    \centering
    \includegraphics[width=0.95\textwidth,height=0.85\textheight,keepaspectratio]{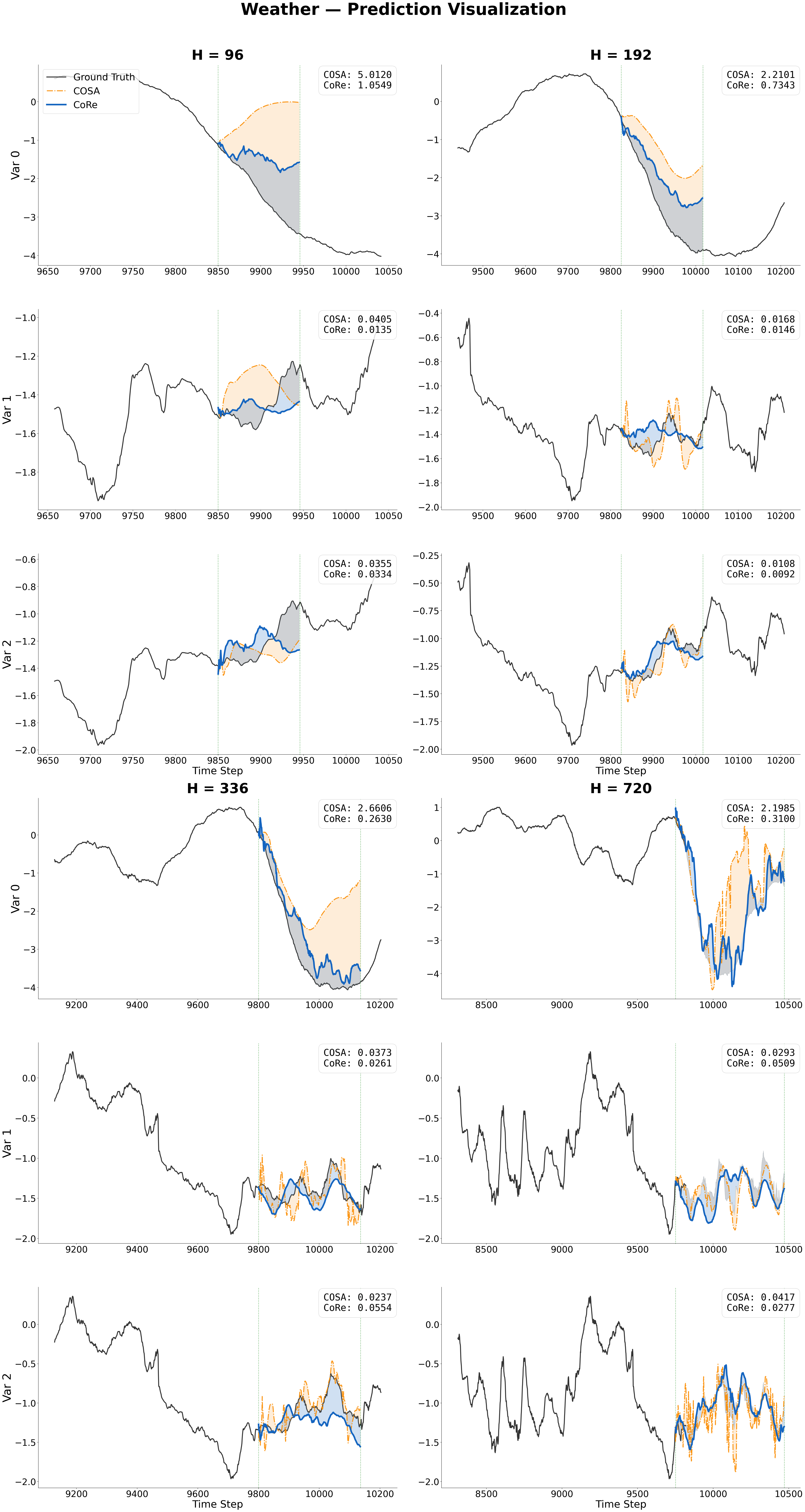}
    \caption{
    Qualitative comparison between COSA and \textsc{CoRe} on Weather
    across four prediction horizons and three selected variates.
    }
    \label{fig:pred_vis_weather}
\end{figure*}

Figures~\ref{fig:pred_vis_exchange_rate} and
\ref{fig:pred_vis_weather} compare COSA and \textsc{CoRe} on
Exchange Rate and Weather.
The differences are visually more pronounced in several
medium-to-long-horizon examples, where \textsc{CoRe} more closely follows
the selected ground-truth trajectories.

On Exchange Rate, the illustrated windows include reductions in
per-window MSE of up to 36\% relative to COSA
(e.g., $H=192$, Var.~6:
$0.0160\rightarrow0.0102$).
These examples are consistent with the aggregate quantitative results in
Table~\ref{tab:main_results_1}.

\subsection{Runtime Breakdown}
\label{app:timing}

Table~\ref{tab:timing_breakdown} reports the runtime decomposition with a DLinear backbone, averaged over six datasets and four horizons. Relative to COSA, the measured per-adaptation-step runtime increases from 6.28\,ms to 6.76\,ms ($+0.48$\,ms; $+7.7\%$). Spectral descriptor computation and SCR account for 1.14\,ms and 1.45\,ms, respectively, while differences in the remaining components
partially offset this additional cost.
The full-stream runtime increases from 11.42\,s to 12.17\,s
($+6.5\%$).

\begin{table}[H]
\centering
\caption{Detailed per-step timing breakdown averaged 
over six datasets and four horizons (DLinear). 
Overhead = \ours{} $-$ COSA; "-" indicates a component 
absent in that method.}
\label{tab:timing_breakdown}
\small
\setlength{\tabcolsep}{6pt}
\begin{tabular}{lrrr}
\toprule
\textbf{Component} & 
\textbf{COSA(ms)} & 
\textbf{\ours  (ms)} & 
\textbf{Overhead(ms)} \\
\midrule
Backbone prediction    & 1.27 & 1.04 & $-0.22$ \\
Spectral FFT             & -  & 1.14 & $+1.14$ \\
SCR forward              & -  & 1.45 & $+1.45$ \\
Loss computation         & 0.84 & 1.00 & $+0.16$ \\
Backward + update        & 4.42 & 4.11 & $-0.31$ \\
\midrule
Per adaptation step      & 6.28 & 6.76 & $+0.48\ (+7.7\%)$ \\
\bottomrule
\end{tabular}
\end{table}

\section{Theoretical Analysis}
\label{app:theory_analysis}

\subsection{Theoretical Connections}
\label{app:theory_connections}

The spectral descriptor summarizes local temporal structure in the
frequency domain.
The Wiener--Khinchin relation connects the power spectrum of a stationary
process to its second-order temporal structure
~\citep{wiener1930generalized,khintchine1934korrelationstheorie}.
We use FFT-derived spectra as empirical descriptors of the finite current
window rather than as estimates of a globally stationary process.

Normalized spectral entropy provides a classical measure of spectral
energy spread
~\citep{powell1979spectral,inouye1991quantification}.
Recent multivariate forecasting work has also used spectral entropy to
guide dependency-aware modeling~\citep{xiong2026seed}.
Together with the band-energy ratios, it provides the compact
four-dimensional descriptor used by the spectral gate.

The structured SCR bottleneck is also consistent with the observed
geometry of the correction matrices.
As shown in Figure~\ref{fig:singular_mode}, the leading singular mode
captures 73\% of correction energy on average.
This observation motivates examining reduced-rank correction interaction
in the high-dimensional experiments.

\subsection{Theoretical Analysis of Interaction Space}
\label{app:theory_formal}

The main text distinguishes prediction-space and correction-space
interaction using generic cross-variate operators.
Here we give a nonlinear characterization of the structural distinction.

Let
\begin{equation}
    \hat{\mathbf Y}^{\mathrm{base}}_t
    =
    \mathbf Y_t
    +
    \mathbf e_t^{\mathrm{base}},
    \label{eq:app_base_error}
\end{equation}
where
$\mathbf e_t^{\mathrm{base}}$
is the frozen-backbone prediction error.

Prediction-space interaction takes the form
\begin{equation}
    \hat{\mathbf Y}^{\mathrm{pred}}_t
    =
    \hat{\mathbf Y}^{\mathrm{base}}_t
    +
    \boldsymbol{\Delta}_t
    +
    \mathcal{M}_{\mathrm{pred}}
    \left(
        \hat{\mathbf Y}^{\mathrm{base}}_t
    \right),
    \label{eq:app_pred_generic}
\end{equation}
while correction-space interaction takes the form
\begin{equation}
    \hat{\mathbf Y}^{\mathrm{corr}}_t
    =
    \hat{\mathbf Y}^{\mathrm{base}}_t
    +
    \boldsymbol{\Delta}_t
    +
    \mathcal{M}_{\mathrm{corr}}
    \left(
        \boldsymbol{\Delta}_t
    \right).
    \label{eq:app_corr_generic}
\end{equation}
The operators may be nonlinear.
For \textsc{CoRe}, the SCR bottleneck and its spectral modulation are
included in $\mathcal{M}_{\mathrm{corr}}$; conditioning on the current input window is
fixed when analyzing the interaction at time $t$.

\subsubsection{Proof of Proposition~\ref{prop:structure}}
\label{app:theory_proof}

The prediction-space interaction output evaluated on the actual backbone
forecast is
\begin{equation}
    \mathcal{M}_{\mathrm{pred}}
    \left(
        \mathbf Y_t+\mathbf e_t^{\mathrm{base}}
    \right).
\end{equation}
To isolate its direct sensitivity to the backbone error, define
\begin{equation}
    \boldsymbol{\Gamma}_t^{\mathrm{pred}}
    :=
    \mathcal{M}_{\mathrm{pred}}
    \left(
        \mathbf Y_t+\mathbf e_t^{\mathrm{base}}
    \right)
    -
    \mathcal{M}_{\mathrm{pred}}
    \left(
        \mathbf Y_t
    \right).
    \label{eq:gamma_pred}
\end{equation}

If $\mathcal{M}_{\mathrm{pred}}$ is continuously differentiable, let
\begin{equation}
    h(s)
    =
    \mathcal{M}_{\mathrm{pred}}
    \left(
        \mathbf Y_t+s\mathbf e_t^{\mathrm{base}}
    \right),
    \qquad
    s\in[0,1].
\end{equation}
By the chain rule,
\begin{equation}
    h'(s)
    =
    J_{\mathcal{M}_{\mathrm{pred}}}
    \left(
        \mathbf Y_t+s\mathbf e_t^{\mathrm{base}}
    \right)
    \left[
        \mathbf e_t^{\mathrm{base}}
    \right],
\end{equation}
where
$J_{\mathcal{M}_{\mathrm{pred}}}(\mathbf A)[\mathbf E]$
denotes the action of the Jacobian of
$\mathcal{M}_{\mathrm{pred}}$ at $\mathbf A$ on perturbation $\mathbf E$.
Applying the fundamental theorem of calculus gives
\begin{equation}
    \boldsymbol{\Gamma}_t^{\mathrm{pred}}
    =
    \int_0^1
    J_{\mathcal{M}_{\mathrm{pred}}}
    \left(
        \mathbf Y_t+s\mathbf e_t^{\mathrm{base}}
    \right)
    \left[
        \mathbf e_t^{\mathrm{base}}
    \right]
    ds.
    \label{eq:nonlinear_error_coupling}
\end{equation}

Equation~\ref{eq:nonlinear_error_coupling} makes explicit that
prediction-space cross-variate interaction is directly sensitive to the
error contained in the frozen-backbone prediction.

In correction space, the interaction operator instead receives
$\boldsymbol{\Delta}_t$:
\begin{equation}
    \mathcal{M}_{\mathrm{corr}}
    \left(
        \boldsymbol{\Delta}_t
    \right).
\end{equation}
Thus,
$\mathbf e_t^{\mathrm{base}}$
does not enter the cross-variate interaction operator directly.
The correction
$\boldsymbol{\Delta}_t$
may itself depend on the backbone prediction through the base adapter;
the structural distinction concerns the quantity on which
cross-variate interaction is explicitly performed.
This proves Proposition~\ref{prop:structure}.

\paragraph{Linear special case.}
For a linear prediction-space interaction applied along the variate
dimension,
\begin{equation}
    \mathcal{M}_{\mathrm{pred}}(\mathbf A)
    =
    \mathbf A\mathbf W_{\mathrm{pred}}^\top,
\end{equation}
Equation~\ref{eq:gamma_pred} reduces to
\begin{equation}
    \boldsymbol{\Gamma}_t^{\mathrm{pred}}
    =
    \mathbf e_t^{\mathrm{base}}
    \mathbf W_{\mathrm{pred}}^\top.
    \label{eq:linear_special_case}
\end{equation}
Hence the explicit coupled-error term used in the linear illustration is
a special case of the nonlinear characterization above.

\subsection{Implication under Delayed Supervision}

The structural distinction above occurs in a delayed-supervision setting.
For a forecast issued at time $t$, the complete target corresponding to
that forecast becomes available only after the prediction horizon.
Consequently, the loss associated with this forecast cannot be used for
a supervised adaptation update before the corresponding target has been
observed.

Prediction-space interaction therefore acts on a backbone prediction containing $\mathbf e_t^{\mathrm{base}}$ before supervision associated with that forecast becomes available. During this feedback window, prediction-space interaction continues to operate on error-containing backbone outputs before the corresponding supervised feedback becomes available. As $H$ increases, the associated supervision arrives later, motivating us to examine whether the empirical consequence of this structural difference varies with the prediction horizon. We evaluate this empirically in Appendix~\ref{app:corrvspred}. Correction-space interaction instead performs its cross-variate operation on the adapter-produced correction while preserving the backbone prediction path.

This argument is a mechanistic motivation, not a guarantee that the performance gap must increase monotonically with $H$. Standard analyses of online learning with delayed feedback likewise contain delay-dependent optimization terms~\citep{langford2009slow}, but we do not use them to claim a performance ordering between the two interaction spaces. The analysis isolates a structural difference between the two interaction spaces; the controlled comparison in Appendix~\ref{app:corrvspred} tests its empirical consequence while holding the refinement architecture fixed.

\section{Limitations and Future Work}
\label{app:limitations}

\paragraph{Short prediction horizons.}
The benefit of correction-space interaction is smaller at $H=96$, with slight degradation in a small number of settings.
Adapting the interaction capacity to the available correction signal may improve short-horizon performance.

\paragraph{Interaction capacity and base-adapter dependence.}
The default choice $r=C$ scales with the number of variates. Although reduced-rank SCR remains effective overall on Electricity and Traffic, its gains are smaller in some high-dimensional settings.
Moreover, \textsc{CoRe} operates on corrections produced by an underlying adapter and therefore depends on the information available in this correction interface.
Results with TAFAS and a standalone MLP show that the interaction principle transfers beyond COSA, while adaptive rank selection and stronger correction interfaces remain directions for future work.

\paragraph{Scope of the analysis and deployment setting.}
Our analysis establishes the structural distinction between prediction- and correction-space interaction rather than an accuracy guarantee.
Experiments follow the standard delayed-supervision protocol and assume backbones that support an additive correction interface.
Extending correction-space interaction to irregular feedback delays and alternative interfaces, including time-series foundation models such as MOIRAI~\citep{woo2024unified} and Chronos~\citep{ansari2024chronos}, is an interesting direction.

\clearpage
\end{document}